# 'Part'ly first among equals: Semantic part-based benchmarking for state-of-the-art object recognition systems

Ravi Kiran Sarvadevabhatla[1], Shanthakumar Venkatraman[2],
Venkatesh Babu R.[1]

[1] Video Analytics Lab, CDS, Indian Institute of Science
Bangalore 560012, INDIA
[2] Indian Institute of Technology - Hyderabad
Hyderabad 502285, INDIA
`ravikiran@grads.cds.iisc.ac.in`,
`shanthakumar792@gmail.com,venky@cds.iisc.ac.in`

**Abstract.** An examination of object recognition challenge leaderboards (ILSVRC, PASCAL-VOC) reveals that the top-performing classifiers typically exhibit small differences amongst themselves in terms of error rate/mAP. To better differentiate the top performers, additional criteria are required. Moreover, the (test) images, on which the performance scores are based, predominantly contain fully visible objects. Therefore, 'harder' test images, mimicking the challenging conditions (e.g. occlusion) in which humans routinely recognize objects, need to be utilized for benchmarking. To address the concerns mentioned above, we make two contributions. *First*, we systematically vary the level of local object-part content, global detail and spatial context in images from PASCAL VOC 2010 to create a new benchmarking dataset dubbed PPSS-12. *Second*, we propose an object-part based benchmarking procedure which quantifies classifiers' robustness to a range of visibility and contextual settings. The benchmarking procedure relies on a semantic similarity measure that naturally addresses potential semantic granularity differences between the category labels in training and test datasets, thus eliminating manual mapping. We use our procedure on the PPSS-12 dataset to benchmark top-performing classifiers trained on the ILSVRC-2012 dataset. Our results show that the proposed benchmarking procedure enables additional differentiation among state-of-the-art object classifiers in terms of their ability to handle missing content and insufficient object detail. Given this capability for additional differentiation, our approach can potentially supplement existing benchmarking procedures used in object recognition challenge leaderboards.

## 1 Introduction

The performance of an object recognition system is typically measured in terms of error rate averaged over the object categories covered. In this respect, various deep-learning based classifiers have shown state-of-the-art performance on large-scale object recognition challenges in recent times. In fact, recognition challenge leaderboards [1,2] typically list classifiers which show minuscule differences in the performance scores, particularly among the top-most performers. Moreover, the scores typically correspond to test images sourced from the same master image set used for training. Using such test images causes the well-documented phenomenon of dataset-bias [3,4,5] to creep into performance scores, thereby presenting a distorted picture of the classifiers' generalization ability. In the face of such observations, an important question arises : how else can these competing systems be differentiated ?

The Holy Grail is, of course, human-like level of performance [6,7]. But, for a recognition system to claim it is within grasping distance of this Grail, the performance criteria can no longer be error rates on mostly fully-visible objects[3] present in biased test images. Instead, we need to design additional and alternative criteria. Also, if we wish to realistically benchmark state-of-the-art classifiers, we require test images which mimic the challenging conditions (e.g. local occlusion, insufficient global context) in which humans routinely recognize objects. To address these concerns, we make the following contributions:

- We systematically vary the level of object-part content, global visibility and spatial context in object images to create a PASCAL-based [8] benchmarking dataset named PPSS-12 (Section 4).
- We propose a novel semantic similarity measure called Contextual Dissimilarity Score (CDS). This measure has been designed to reflect a classifier's ability to predict the target category in a semantically meaningful manner across varying visibility and contextual settings (Section 5).
- We use our measure CDS and the PPSS-12 dataset to benchmark the top-performing object recognition classifiers trained on the ILSVRC-2012 dataset. The results (Section 6) show that our benchmarking procedure enables additional differentiation between the top-performers on the basis of their ability to handle missing content and incomplete object detail.

## 2   Overview of our approach

Figure 1 provides an overview of our approach. In the text that follows, circled numbers correspond to various data items and processing stages of our approach, as marked in Figure 1.

We benchmark the top-performing [9] object classifiers trained on the 1000-class ILSVRC-2012 dataset – Alexnet [10], VGG-19 [11], NiN [12] ,GoogLeNet [13]. For benchmarking purposes, we first create 'PASCAL Parts Simplified (PPS)-12' – a modified, 12-category image subset (②) of PASCAL-parts [14] which in turn is a database of object images with semantic-part annotations (Section 3).

For each image $I$ (①) in PPS-12 containing a reference object, we systematically vary the object's level of global visibility and its spatial context in terms of semantic object-parts (③) to create an associated sequence $S_I$ of images (④) (Section 4). The collection of all such sequences comprise our benchmarking dataset 'PASCAL Parts Simplified Sequences (PPSS)-12' (⑤).

---
[3] Anecdotally, this is the case in most object-recognition datasets.

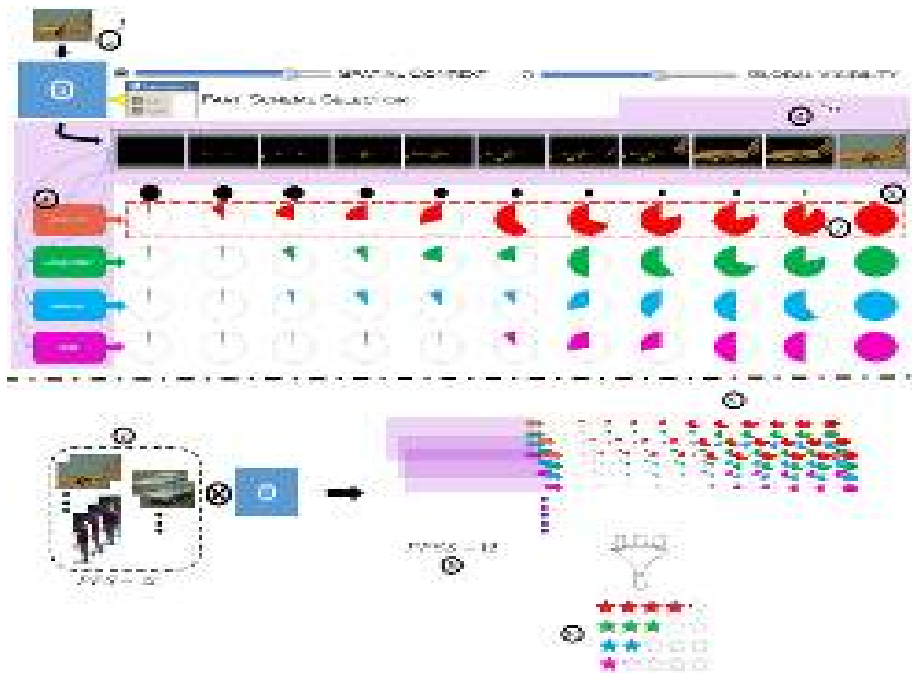

**Fig. 1.** A graphic overview of our approach (Section 2). For each image $I$ (①) in PPS-12, we systematically vary the object's global visibility / spatial context in terms of its parts (③) to create sequence $S_I$ of images (④). The main processing block is shown shaded in purple background above the black dash-dotted line. ⑤ refers to collection of such blocks, each of which contains the image sequences that form our benchmarking dataset $PPSS - 12$. The ⊗ in the lower half indicates that the various global visibility schemes/spatial context schemes are applied to the base dataset $PPS - 12$ to create the sequences which form $PPSS - 12$. For each sequence image, the degree of semantic similarity between its ground-truth label and that predicted by a classifier (⑥) is depicted as the proportion to which the corresponding circle underneath the image is filled (⑦). The similarity value of each sequence image is associated with a position-in-sequence based weighing factor depicted by the relative size of the black filled circle (⑧). The weighted similarities for the sequences in PPSS-12 are analyzed (⑨) to benchmark the classifiers (⑩). Best viewed in color.

Having obtained the sequences, we first fix a classifier (⑥). For each image in the sequence $S_I$, we determine the normalized semantic similarity (Section 5) between the classifier-predicted label and ground-truth label[4]. We associate the similarity score of each image in the sequence with a normalized weight factor such that the earlier the relative location of the image within the sequence, the greater its weight[5]. We compute the weighted sum of similarity scores and normalize them by the sequence length to obtain a similarity-measuring score $s$. To obtain a measure similar in interpretation as error rate (i.e. lower the better), we subtract $s$ from 1 to arrive at the final classifier-specific dissimilarity score which we term Contextual Dissimilarity Score $CDS_I$ for image $I$ (Section 5). By construction, early images of the sequence $S_I$ contain relatively smaller amount of evidence for the reference object (see Figures 2,3). Therefore, the larger the semantic similarity between classifier predictions and ground-truth for the initial images of sequence $S_I$, the greater the ability of the classifier to predict the target category in a semantically meaningful manner in challenging visibility and contextual settings and demonstrate human-like performance. This ability is numerically characterized by a low average CDS for the classifier. We gather statistics on CDS in the image sequences (ⓖ) on a per-image and per-classifier basis, across object categories. These statistics enable us to benchmark the classifiers as desired (ⓗ) (Section 6).

---
[4] The degree of similarity is depicted as the proportion to which the circle underneath each sequence image is filled - see ⑦ in Figure 1.

At this juncture, the reader might be inclined to question aspects of our semantic part-based benchmarking approach. We discuss the reasons and consequences of our choices in Section 7. For now, we move on to describe our preprocessing of the PASCAL-parts dataset.

## 3   Data preprocessing

For the purpose of benchmarking, we start with a subset of PASCAL-parts [14], a 20-category object dataset containing named-part (semantic) annotations[6]. From these, we shortlist 12 categories (`aeroplane, bicycle, bus, car, cat, cow, dog, horse, person, sheep, train`) using the following criteria: (1) presence of at least two annotated parts (2) ease of annotating additional parts as required. We used object bounding box annotations from PASCAL-parts to obtain the cropped object images.

The part labeling scheme in PASCAL-parts contains labels on the basis of object orientation (left-facing, right-facing etc.) and intra-object location (leg-top, leg-bottom etc.). We simplified this scheme by ignoring such factors (i.e. orientation, location). In addition, certain crucial object parts have not been annotated in PASCAL-parts dataset. Therefore, we modified the PASCAL-parts' labeling scheme by adding such parts to the annotation scheme. We used an in-house labeling tool to obtain annotations for these additional crucial parts.

In the end, we obtain an image set with 1850 object images across 12 object categories. We refer to this subset of PASCAL-parts with simplified/modified part annotations as PPS-12. Each object image in PPS-12 (See ① in Figure 1) forms the basis for the construction of the image sequences in our benchmarking dataset PPSS-12[7]. Next, we describe how these image sequences are actually constructed.

---

[5] The weighing factor is depicted by the relative size of the black filled circle - see ⓢ.

[6] In this paper, we refer to named parts interchangeably with semantic parts.

[7] Our dataset is available at http://val.serc.iisc.ernet.in/pbbm/

## 4   Image Sequence Construction

For each object image $I$ in PPS-12, we construct a sequence $\hat{S}_I$ of images (See ④ in Figure 1). This sequence typically begins with only one semantic-part of the object in the image. The remaining object parts are successively added using a pre-defined 'part ordering scheme' (Section 4.1) to form the rest of the sequence. For each image in the sequence $\hat{S}_I$, a pre-defined 'content scheme' is applied to obtain the new sequence $\mathcal{S}_\mathcal{I}$. The content scheme controls either the amount of spatial context or level of object detail within the sequence (Section 4.2). Also, $\mathcal{S}_\mathcal{I}$ is constructed such that the final image in the sequence always coincides with the object image $I$ which serves as the basis for the sequence construction in the first place. The collection of all such sequences constitutes our benchmarking dataset PPSS-12.

For the remainder of the section, we first describe the different object-part ordering schemes used to create image sequences. We subsequently describe the object content schemes (Section 4.2) activated during image sequence creation.

### 4.1   Part-ordering schemes

The part-ordering schemes essentially produce an ordered list of parts for each object category based on the scheme's criteria. During the image sequence creation (Section 4.2), this list forms the basis for incremental addition of object content.

In a recent work, Li et al. [15] augmented 850 images from PASCAL 2010 dataset with eye-fixation information to create their dataset PASCAL-S. In this dataset, each image is associated with a set of eye-fixation sequences. Each eye-fixation sequence, in turn, corresponds to spatial locations fixated upon by a human subject's eyes when shown the image. The PASCAL-parts derivative dataset we have created, PPS-12, is also derived from PASCAL 2010. We first identify images common to PASCAL-S and PPS-12. For these images, we analyze the density of fixation locations (from PASCAL-S) with respect to object part boundaries (from PPS-12). For each object category, we sort the parts in the decreasing order of fixation density to obtain the part-ordering scheme. In doing so, we implicitly make the assumption that fixation density is correlated with relative importance of image content, a phenomenon repeatedly observed in eye-fixation based image saliency studies [16]. We explored four variations in determining per-part fixation density resulting in four eye fixation-based part ordering schemes. We describe these schemes next.

Let $F^I(k) = \{f_j(k) = (x_j, y_j)\}, j = 1, 2, \ldots N_k$ denote the $k$-th eye fixation sequence (out of the $N$ sequences from PASCAL-S) for an image $I \in$ PPS-12 containing an object from category $\mathbb{C}$. Here, $(x_j, y_j)$ corresponds to the spatial location of the eye-fixation. Let $p_j(k), 1 \leq j \leq N_k$ denote the part within whose spatial boundary fixation $f_j(k)$ lies. $N_k$ denotes the number of fixations in the $k$-th fixation sequence. Let $\mathcal{P}^\mathbb{C}$ denote the set of parts associated with category $\mathbb{C}$.

**Unnormalized sequence position scheme ($E_{US}$):** Under this scheme, we assign part importance based on the total number of fixations within a part $P$'s boundary. However, we also weigh each fixation by the relative position of the fixation within its original sequence. The part scheme factor for part $P$ (of category $\mathbb{C}$) in image $I$ is computed as :

$$E_{US}(P) = \frac{\displaystyle\sum_{k=1}^{N}\sum_{j=1}^{N_k} r_j(k)1(P = p_j(k))}{\displaystyle\sum_{Q \in \mathcal{P}^\mathbb{C}}\sum_{k=1}^{N}\sum_{j=1}^{N_k} r_j(k)1(Q = p_j(k))} \tag{1}$$

$$\text{where } r_j(k) = \frac{N_k - j + 1}{N_k} \tag{2}$$

Here, the factor $r_j(k)$ captures the intuition that the earlier a fixation's location within its fixation sequence, the greater the prominence of the part within whose contour it falls. In the above equation and those that follow, 1 denotes the indicator function.

**Unnormalized part count scheme ($E_U$):** This scheme is similar to $E_{US}$ except that all fixations are considered equally important, i.e. $r_j(k) = 1$. Under this scheme, we simply count the number of fixations that lie within a part $P$'s boundary as a measure of part importance.

**Part area normalized schemes ($E_A$ and $E_{AS}$):** In the $E_U$ scheme, a part can have a large importance score merely because it covers a larger portion of the image. Following Kiwon et al. [17], we normalize for part areas and construct part-area normalized versions of $E_U$ and $E_{US}$ as follows :

$$E_A(P) = \frac{E_U(P)}{A(P)} \tag{3}$$

$$\text{where } A(P) = \frac{\text{Area of part } P}{\text{Area of the object within } I} \tag{4}$$

$$E_{AS}(P) = \frac{E_{US}(P)}{A(P)} \tag{5}$$

In all the part schemes mentioned above, we sum the part-importance factor (Equations 1, 3, 5) for each part across all images of the category and sum-normalize to obtain the probability distribution of relative part importance. We obtain the final part scheme by listing the parts in the order of decreasing probability.

### 4.2   Content (Global object visibility and object context) schemes

Having described the part schemes, we next describe how systematic variations in context and global object visibility are introduced into object images (See ③ in Figure 1). We refer to the schemes which control image variations in these aspects (global object visibility, object context) as 'content schemes'. The content schemes essentially control the level of detail and the manner in which this detail is added as we progress across a part-image sequence. From a benchmarking point of view, these schemes are designed to evaluate the extent to which lack of content or the ability to exploit existing context affects a classifier's performance.

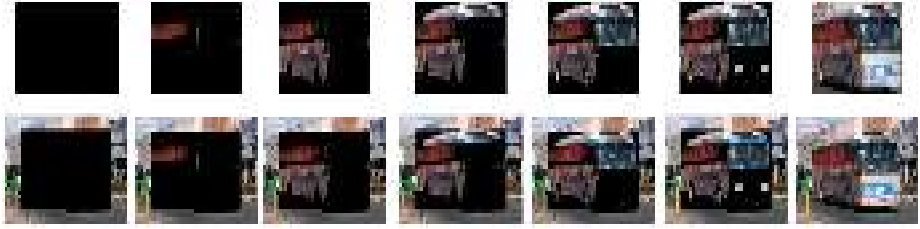

**Fig. 2.** Part sequences corresponding to change in 'object context' of a `bus` image. The top row corresponds to 'intra-object' context scheme while the bottom row corresponds to 'intra-and-neighborhood' context. Notice the road-like surroundings providing the context (surrounding the object) in the bottom row.

**Object context scheme:** Two variations exist within the object context scheme. In the first variation ('Intra-object context'), the images of the part-image sequence contain no contextual information other than that arising from incremental addition of object's parts. To ensure this, the object's parts are added to a completely black canvas. Figure 2 (top row) shows an example. In the second variation ('Intra-and-Neighborhood context'), the image content immediately surrounding the object is retained to provide neighborhood context. However, parts are incrementally added within a blacked-out bounding box enclosing the object. Figure 2 (bottom row) shows an example.

**Global object visibility scheme:** Unlike the context-based schemes mentioned above, the visibility schemes additionally have access to global context from the entire image in a gist-like manner, including that from parts not yet added. Figure 3 shows example sequences for this content scheme. In this scheme, the entire image's data is present, albeit at a low level of detail to begin with. As each part of the object is added, the part comes into focus. The net effect is a blurring of the image relative to already added parts. This scheme is inspired by the manner in which level of detail falls relative to the location fixated upon by a human eye. To achieve this fall-off effect, we utilize the visual-field simulation of Perry and Geisler [18]. By varying the parameters of the visual-field simulation, the level of detail in the immediate vicinity of each added part can be changed. For the purposes of our evaluation, we utilize two parameter settings which result in two variations of the global object visibility scheme which we refer to 'low level of detail' (top row of Figure 3) and 'higher level of detail' (bottom row).

In the next section, we describe how the PPSS-12 sequences created by applying the content schemes to images in PPS-12 help determine the 'Contextual

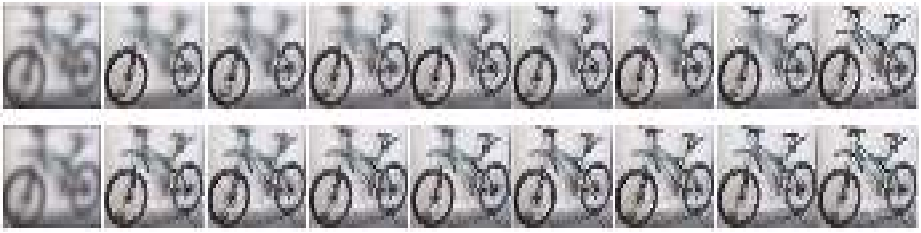

**Fig. 3.** Part sequences corresponding to change in global object visibility of a `bicycle` image. We refer to the two variations of the visibility level as 'low level of detail' (top row) and 'higher level of detail' (bottom row).

Dissimilarity Score' (CDS) for a fixed classifier. Later on (Section 6), we shall see how the process of determining CDS lets us benchmark the classifiers.

## 5    Determining Contextual Dissimilarity Score ($CDS$)

For a given image $I$ from our PPS-12 dataset (①, Fig. 1), we first choose a part-ordering scheme (Section 4.1) and generate a sequence of images $\mathcal{S}_I$ according to this scheme. We then choose an object content scheme (Section 4.2) and apply it to each of the images in the sequence (③, Fig. 1). With the content-scheme applied sequence at hand (④, Fig. 1), we are ready to determine the Contextual Dissimilarity Score $CDS_I$ for image $I$.

Let $\mathcal{S}_I = \{S_1, S_2, \ldots, S_N\}$ represent our aforementioned image sequence from PPSS-12. Note that by construction, sequence image $S_N$ corresponds to given image $I$. Since our analysis is on a per-classifier basis, let us fix the classifier $\mathbb{C}$ (⑥, Fig. 1). Each image in sequence $S_j \in \mathcal{S}_I$ is input to the classifier to obtain the corresponding class prediction label $c_j$. Suppose the ground-truth label for $S_j$ is $g_j$. In our case, $c_j$ and $g_j$ are drawn from two different label spaces (Imagenet-based and PASCAL-based) with varying levels of semantic granularity and therefore, an exact literal match may not be possible. Therefore, we utilize a semantic similarity measure $\mathcal{M}$ which provides a $[0, 1]$-normalized score $x_j$ reflecting the semantic similarity between $c_j$ and $g_j$ (i.e. $x_j = \mathcal{M}(c_j, g_j)$). Thus, we obtain a sequence of normalized scores $\mathcal{X}_I = \{x_1, x_2, \ldots, x_N\}$[8].

Also note that by construction, early images of the sequences contain relatively smaller amount of evidence for the reference object (see Figures 2, 3). Therefore, the higher the similarity score in the initial parts of the sequence, the greater the ability of the classifier to perform well in challenging conditions and demonstrate human-like performance. To characterize this notion, each similarity score $x_j$ in the sequence in associated with a weight factor $w_j = \frac{N-j+1}{N}$ such that the earlier the location, the greater its weight (⑧, Fig. 1)). We then compute $NWSS$ - the normalized weighted sum of similarity scores. To obtain a measure similar in interpretation as error rate (i.e. lower the better), we subtract $NWSS$ from 1 to arrive at the final Contextual Dissimilarity Score for image $I$ ($CDS_I$).

---

[8] The colored rows containing the pie-like shapes in Figure 1 correspond to such similarity scores.

$$CDS_I = 1 - \frac{\sum_{j=1}^{N} x_j w_j}{\sum_{j=1}^{N} w_j} \qquad (6)$$

The resulting $CDS_I$ is an indicator of the part-level and contextual content required by classifier $\mathbb{C}$ to recognize the object in image $I$. Therefore, obtaining a relatively smaller 'average CDS' when $CDS_I$ are averaged across part-schemes, context-schemes and object categories indicates the ability of a classifier to perform well in spite of missing or poorly detailed object information.

## 6  Experimental Analysis

| Classifier | GoogLeNet | VGG-19 | NiN | Alexnet |
|---|---|---|---|---|
| Median error rate | 0.24 | 0.25 | 0.32 | 0.35 |

**Table 1.** Cross-dataset error-rates: Performance of the ILSVRC trained classifiers on our PASCAL-based PPS-12 dataset using manual mappings across the two datasets.

As a preliminary experiment, we computed the median top-1 error-rate of each classifier (ILSVRC-trained) on our PPS-12 (derived from PASCAL) images. Given the inherent dataset bias commonly present in recognition challenge datasets [3], the error-rates are higher unlike the low, barely distinguishable top-1 rates typically encountered on recognition challenge leaderboards (See Table 1). This result should not be surprising. Instead, it merely serves to reinforce the importance of cross-dataset validation in obtaining a fair assessment of classifiers' generalization capabilities [19].

As part of the main benchmarking process, we determine the CDS (Section 5) for all possible combinations of classifiers, part-ordering schemes (Section 4.1) and object content schemes (Section 4.2). This sets the stage for examining the effect of these schemes on the overall benchmarking process. For the similarity measure $\mathcal{M}$ between the category labels (Section 5), we used Wu-Palmer similarity measure [20]. This measure calculates relatedness of two words using a graph-distance based method applied to WordNet [21], a standard English lexical database containing groupings of cognitively similar concepts and their interrelationships.

| Scheme | Context based | | Global visibility based | |
|--------|---------------|--------------------------------|---------------------|-----------------------|
| | Intra-object | Intra-object and neighborhood | Low level of detail | Higher level of detail |
| ALEXNET | 0.4499 | 0.4470 (0.6 %) | 0.4450 | 0.3803 (14.54 %) |
| GOOGLENET | 0.5264 | 0.4319 (**17.95 %**) | 0.4544 | 0.3490 (**30.20 %**) |
| NIN | 0.4788 | 0.4492 (6.18 %) | 0.4689 | 0.3882 (17.20 %) |
| VGG-19 | **0.4136** | **0.4147** (-0.27 %) | **0.3628** | **0.2880** (20.62 %) |

**Table 2.** Benchmarking classifiers : Average median CDS across categories for different context schemes. The best CDS score for each content scheme is shown in bold. The bracketed percentages in column 2 indicate the improvement in CDS over column 1 with addition of context. The ones in column 4 indicate the improvement over column 3 when level of detail is increased. The best percentage improvement is also shown in bold. Note that smaller the CDS, better the performance.

## 6.1 Benchmarking classifiers across object content schemes

In the discussion that follows, it is important to remember that smaller the CDS, better the classifier's performance.

**'Intra-object' context:** For the first set of experiments, we analyze CDS for the 'intra-object' context scenario. This scenario consists of object images without any of the surrounding context except that arising out of the object's parts themselves (see top row of Figure 2) and is perhaps the most challenging scenario for a classifier. On the other hand, it is also the most appropriate since the image content is precisely confined only to the object.

Fixing the content scheme to 'intra-object', for each classifier and for each category, we compute the median CDS. We do this initially for each part-ordering scheme and subsequently average the median scores over the schemes to obtain category-wise CDS. These category-wise scores are, in turn, averaged to obtain the CDS for each classifier. The results on a per-classifier basis can be seen in the first column ('Intra-object') of Table 2 . As expected, the median scores are relatively high regardless of classifier.

**'Intra-object and neighborhood' context:** We repeat the previous experiment with the content-scheme now being 'intra-object and neighborhood'. In addition to object parts, contextual information from the immediate surroundings is additionally available in this scheme (see bottom row of Figure 2). We hypothesized that such information would improve performance and that is indeed the case (see second column ('Intra-object and neighborhood') of Table 2).

**Global object visibility:** Next, we examine the impact of visibility-based content schemes (Section 4.2). As mentioned before, these schemes, unlike the intra-object and/or neighborhood context ones, have additional access to global context from the entire image in a gist-like manner, including that from parts not yet brought into focus (See Figure 3). Therefore, the performance of the classifier for these schemes conveys the extent to which it utilizes the global context.

Keeping the classifier fixed and content-scheme as 'low-detail', for each category, we compute the median CDS for each part scheme and average them across part schemes to obtain category-wise CDS. These are averaged in turn to obtain the CDS for the classifier. As the results in Table 2 (third column) suggest, the presence of global information, even at a low level of detail and even with minimal object-specific information, is still powerful enough to improve performance, as evidenced by the lower CDS. Increase in the level of detail (i.e. lower level of blurring) causes the results to be on predictable lines, with the overall average median CDS trending downwards (See last column of Table 2).

### 6.2   Overall performance and additional experiments

Examining the results in Table 2, it is evident that VGG-19 achieves the best performance (lowest average CDS) in general. More importantly, Table 2 also shows that our benchmarking procedure contrasts the performance of almost equally well-performing classifiers (GOOGLENET, VGG-19) better than the traditional accuracy-based counterparts — the CDS-based benchmarking values are generally further apart compared to the accuracy scores (compare Tables 1, 2).

To determine which classifier exploits addition of object neighborhood-based context the most, we compute the percentage improvement in average CDS over the 'object only' (i.e. no neighborhood context) setting (Table 2,first column). As the bracketed numbers in second column of Table 2 show, GOOGLENET's performance improves the most. GOOGLENET also best exploits the increase in level of detail (fourth column of Table 2). We believe these results stem from the 'inception-style' mechanism GOOGLENET [11] uses to capture context.

To obtain a category-level perspective on the benchmarking performance, we determine the classifier that produces the lowest CDS most frequently across all combinations of part schemes and content schemes. The entries in Table 3 (top row) merely endorse the results seen earlier – VGG-19 is the best performer in general. At the other end, NIN and surprisingly (for a couple of categories), GOOGLENET have relatively higher CDS (bottom row of Table 3).

### 6.3   Relationship between CDS and (traditional) error measures

To verify that our CDS measure provides additional information beyond the traditional top-1 error measure, we computed the correlation between CDS and the top-1 error rates across all the classifiers. For this, we determined the median error-rate and median CDS for each content scheme by averaging across the respective measures across part schemes and classifiers. Thus, we obtain two vectors, one for median error-rate and the other for median CDS. The correlation between these two vectors was found to be close to 0 (Pearson $\rho = 0.0227, p = 0.98$ and Spearman $\rho = 0, p = 1$). This result indicates that CDS measures an aspect of classifier performance distinct from the traditional top-1 measure.

| Classifier | airplane | bicycle | bird | bus | car | cat | cow | dog | horse | person | sheep | train |
|---|---|---|---|---|---|---|---|---|---|---|---|---|
| Lowest-cds | GOOGLENET | VGG-19 | VGG-19 | VGG-19 | VGG-19 | VGG-19 | VGG-19 | VGG-19 | VGG-19 | VGG-19 | ALEXNET | VGG-19 |
| Highest-cds | NiN | NiN | NiN | NiN | NiN | NiN | NiN | GOOGLENET | NiN | GOOGLENET | GOOGLENET | GOOGLENET |

**Table 3.** Category-wise best and worst performers (in terms of CDS) aggregated across part and content schemes.

## 7 Discussion and Related Work

Having presented the experiments and analysis, we now examine some of the design decisions and forces at play in our work.

Our benchmarking procedure relies crucially on semantic object part-based image sequences. Using 'named' semantic-parts ensures that all images of a category are treated *consistently*. This advantage is lost when we use purely statistically generated, unnamed , region-based part models[9] [22]. On a related note, Taylor et al. [23] suggest that humans tap into generic concepts of objects, including linguistic propositions (e.g. named object-parts) while analyzing a scene. Furthermore, studies by Palmer [24] have shown that when parts correspond to a 'good' segmentation of a figure (e.g. object-part contours), the speed and accuracy of responses related to queries on figure attributes improves significantly. These observations further lend support for our use of semantic named object-parts. The burdensome aspect of semantic-part annotation does limit the number of categories benchmarked. However, recent trends seem to suggest the possibility of large, richly detailed datasets [25] and multi-task recognition frameworks [26] which can potentially offset this burden.

Our choice of part-importance order (Section 4.1) offers a future opportunity to explore connections between eye-fixation based saliency properties of a partial content image and its recognizability. We wish to point out that our part ordering schemes are not exhaustive – any other principled part-ordering scheme may also be utilized for additional hold-out style benchmarking. In this respect, it is interesting to note that Taylor et al. [23] suggest a list specifying a Bayesian prior on possible object attributes (including semantic-parts) to characterize objects and related concepts.

The images from the sequences we used to compute the CDS are artificial in construction and one might argue that they are too structured and therefore, an imperfect representation of the object occlusion scenarios typically seen in real photos. An alternative could be to utilize realistic data wherein the extent and the manner in which the target object is occluded can be precisely quantified. This, in itself, is an extremely challenging task although newer datasets with depth ordering and occlusion level specified as part of annotations [27] may compensate to some extent[10]. The advantage of our constructs is that they let us quantify the global object visibility *consistently* – for a given location in the part importance order, the *same* part is missing in all the image sequences. Moreover, as the results indicate, state-of-the-art classifiers can still utilize available information effectively in spite of the artificial nature of sequence images.

---

[9] The lack of consistency also holds true for area-based approaches (e.g. systematically decreasing the percentage of object area occluded by a fixed percent).

[10] The dataset was not publicly available at the time of our publication.

Our benchmarking measure relies on a semantic measure of dissimilarity between the predicted label and ground-truth label. The deeper implication of our choice of similarity measure is that the median CDS for each classifier reflects the general ability of the classifier to utilize the semantics of the image to produce semantically meaningful predictions. We initially considered an alternative scheme : a more traditional 'hard' $0-1$ binary prediction in place of 'soft' semantic similarity. However, this approach requires a manual, subjectively grouped, many-to-one mapping between predicted-label set (Imagenet) and ground-truth label set (PASCAL).

On a deeper level, our overall approach reveals aspects of the object recognition task that each of the top-performers address better than the rest. As already pointed out, while VGG-19 is the top-performer in general, GoogLeNet is better (in percentage terms) at exploiting context from an object's immediate surroundings (Section 6.2). Therefore, while our benchmarking procedure is useful to differentiate classifiers, it can also be used to characterize the extent to which contextual and visibility factors are addressed by a classifier on a stand-alone basis. Such characterization can help classifier designers tweak their architectures and help improve the classifier's capabilities. In addition, as Table 2 shows, our benchmarking procedure contrasts the performance of almost equally well-performing classifiers (GoogLeNet, VGG-19) reasonably better than the traditional approach (Table 1). In addition, the moderately high CDS scores (Table 2) suggests that top of the line classifiers of current day are yet to perform well on images which mimic the challenging conditions (e.g. occlusion) in which humans routinely recognize objects. To confirm that humans recognize the objects much more robustly than machine classifiers, we performed a rudimentary user study in which we asked human subjects to recognize the PPSS-12 sequence images. We found that human CDS values were indeed disproportionately low compared to the classifiers[11]. Finally, we also wish to point out that our benchmarking procedure is by no means complete - a gamut of additional transformations (e.g. rotation) and their combinations can be applied to create additional image sequences and benchmark them using the approach described in our work.

Typically, the state-of-the-art results reported on recognition leaderboards [2,1] and literature [28,29,6,30] correspond to ensemble models. However, the corresponding pre-trained models were not always available. To keep the benchmarking consistent, we utilized readily available, pre-trained, non-ensemble baseline models [9].

**Related Work:** One class of quantitative approaches which supplement the usual mAP/error-rate essentially use variations of the traditional measures or tend to be derived from them [31,32]. These additional measures (e.g. Area-Under-the-Curve(AUC), precision, recall) may provide additional differentiation between classifiers but unlike our work, do not provide insight into semantic aspects of data which affect classification. Somewhat similar to our approach, Aghazadeh et al. [33] propose measures which quantify properties of training data (class bias, intra-class variation) and compare classifier performance on the basis of such measures for an object detection problem. However, their formulation involves comparison of features across image pairs whereas our measure is based on per-image statistics aggregated over a category. Hoiem et al. [34] characterize the effects of challenging extrinsic factors (e.g. occlusion, viewpoint) and intrinsic factors (e.g. aspect ratio, part visibility) for the object detection problem and suggest the factors most likely to impact performance. Analyzing user study data for downsampled versions of $256 \times 256$ images, Torralba et al. [35] examine the effect of image resolution for scene and object recognition. However, their study is focused on human subject performance.

---

[11] In fact, the humans were able to correctly recognize the category at extremely early stages of the sequence – the highest median score across content schemes was $0.20(\pm0.06)$ while the lowest was 0.

## 8   Conclusion

In this paper, we have demonstrated a semantic part-based procedure for benchmarking state-of-the-art classifiers. The benchmarking procedure relies on a semantic similarity measure that naturally addresses potential granularity differences between the category names in training and test datasets, thus eliminating laborious and subjective manual mapping. The measures we propose provide additional insights into the classifiers' ability to handle various degrees of object detail and missing object information *à la* humans. In our particular case, the benchmarking procedure enables performance evaluation of the ILSVRC-trained classifiers for test images sourced from an different dataset (PASCAL). Given this capability for additional differentiation, our benchmarking procedure can supplement existing procedures used in object recognition leaderboards. In addition, our benchmarking procedure and dataset are potentially useful for classifier designers on a standalone basis to analyze their classifier's ability to handle missing content and incomplete object detail.

The top performers in our benchmarking study do not explicitly consider object parts (semantic or otherwise) nor do they attempt to model occlusions. However, architectures which are "part-aware" [36] and explicitly contain compensatory mechanisms for occlusion [37] hold great potential not only for our benchmarking procedure, but for the broader area of object recognition as well [7].

Please visit our project webpage http://val.serc.iisc.ernet.in/pbbm/ for additional and up-to-date information.

**Acknowledgement:** We wish to acknowledge NVIDIA for their generous grant of the GPU. This work was partially supported by Science and Engineering Research Board (SERB), Department of Science and Technology (DST), Govt. of India (Proj No. SB/S3/EECE/0127/2015) and Defence Research and Development Organization (DRDO), Govt. of India.

# Supplementary-A: Additional details on data collection, image sequence creation and semantic-part analysis

Ravi Kiran Sarvadevabhatla, Shanthakumar Venkatraman, Venkatesh Babu R.

Video Analytics Lab, CDS, Indian Institute of Science
Bangalore 560012, INDIA
`ravikiran@grads.cds.iisc.ac.in`

## 1  Data pre-processing

**Preliminary selection of benchmarking categories:** We benchmark the top-performing [1] object classifiers trained on the 1000-class ILSVRC-2012 dataset – Alexnet [2], VGG-19 [3], NiN [4] and GoogLeNet [5]. To construct our test data, we utilize a subset of PASCAL-parts [6], a 20-category object dataset containing semantic-part annotations. From these, we shortlist 12 categories (`aeroplane`, `bicycle`, `bus`, `car`, `cat`, `cow`, `dog`, `horse`, `person`, `sheep`, `train`) using the following criteria: (1) presence of at least two annotated parts (2) ease of annotating additional parts as required (3) overlap with categories of the Eitz et al. [7]'s freehand sketch dataset. The last criterion aims to serve a lateral project exploring connections between object images and their hand-drawn versions. However, we shall not delve into the details here.

**Cropping and resizing objects from their original images:** The object images from the 12 categories are usually present as part of a scene. Since our focus is on the object for the purpose of our benchmarking, we crop all objects belonging to the set of 12 target categories mentioned above and group them according to their respective categories. However, in addition to a tight crop (restricted only to the bounding box of the object), we also crop the images to include an extent of 18% of each image's dimensions to include the surrounding context[1]. Thus, we retain this latter version of the object to examine the manner in which contextual information is utilized by a given classifier and across classifiers in general.

## 2  Annotation

The part labeling scheme in PASCAL-parts contains labels on the basis of object orientation (left-facing, right-facing etc.) and intra-object location (leg-top, leg-bottom etc.). We simplify this scheme by ignoring these factors. In addition, certain crucial object parts have not been annotated in PASCAL-parts dataset.

---

[1] In fact, we select only those object images whose image dimensions are 18% more in at-least one of vertical or horizontal dimension.

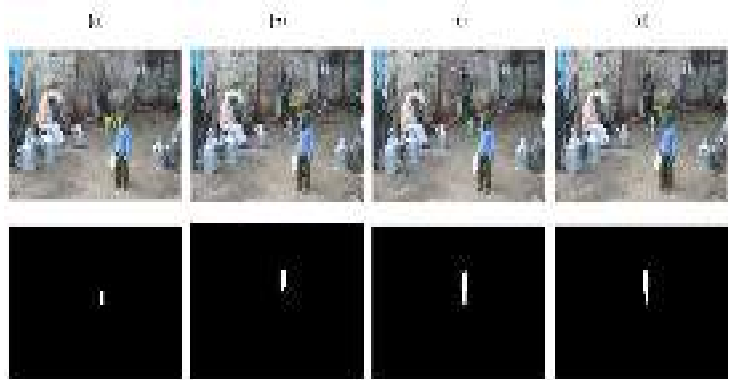

**Fig. 1.** The top row contains annotation contours (outlined in green) for 'rleg' – right leg of a person. The bottom row contains the corresponding part masks. Columns (a) and (b) show annotations to be merged. Column (c) shows the effect of naively merging the annotations. Column (d) shows the result of alpha-shape based merging of annotations.

We add these missing parts using an annotation tool developed in-house. Overall, we refer to our labeling scheme as SPL (Simplified Part Labeling). Specifically, we use the following guidelines for annotating the object images :

1. If an object part in PASCAL-part labeling scheme is known by a different alias under SPL scheme, we simply rename the part by the alias.
2. If an object part is split up as multiple annotations, we merge the multiple annotations into a compact, single annotation by determining the $\alpha$-shape of the participating annotation regions (See Figure 1). Please refer to Section 2.1 for details of the $\alpha$-shape based annotation merging procedure.
3. If a part existing in the SPL scheme is visible in the image but has not been annotated, we shortlist the part for manual annotation.
4. After addressing the above guidelines, it is possible that certain areas of the object remain unlabeled. These are annotated with the generic part label 'unlabeled-rest' (see Figure 2).

### 2.1   $\alpha$ - shape method for merging annotations

As part of simplifying annotations from an existing dataset, we sometimes need to merge two annotations. For instance, we might have to merge upper and lower portions of a leg of an animal into a single, closed contour annotation. To ensure that the merged annotation is as compact as possible, we use the method of $\alpha$-shapes. To begin with, let us assume that we have 2-D spatial masks corresponding to annotations we wish to merge. The objective is to merge these masks into a final mask which excludes unnecessary image regions as much as possible.

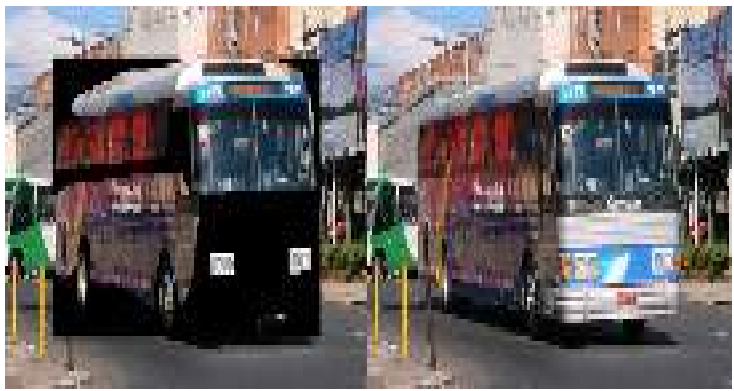

**Fig. 2.** The object parts which are left unlabeled by annotators and object background (black portions on left) are assigned the label 'unlabeled-rest'. These are added last to get back the original reference image (right) when the image sequence is created – refer to Figure 2 of the main paper.

Many methods exist to compute a compact spatial boundary of a given set of 2-D points. One possible baseline approach would be to find the convex-hull of union of regions. A potential drawback of the latter (i.e. convex hull) is that the convexity constraint can cause the boundary to include large amounts of empty space, e.g. when the point distribution consists of island-like sub-clusters connected by slender isthmuses or when the point distribution resembles an annular ring.

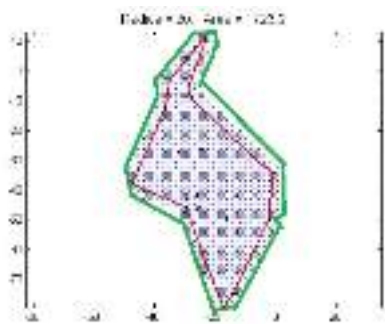

**Fig. 3.** Boundary estimation using $\alpha$-shapes for the region denoted by blue crosses. The initial boundary estimate is shown as a red line – this omits many points belonging to the region. The final boundary obtained using iterative refinement of the original boundary is shown as the green line.

An alternate method, termed α-shapes, relaxes the convexity requirement and instead constructs a non-convex boundary [8] (see Figure 3). One of the parameters, conventionally referred to as α-radius determines the degree of convexity in the estimated boundary. We determine the optimum value of α-radius using iterative refinement so that a minimally enclosing compact boundary is obtained.

## 2.2   Annotation tool and guidelines

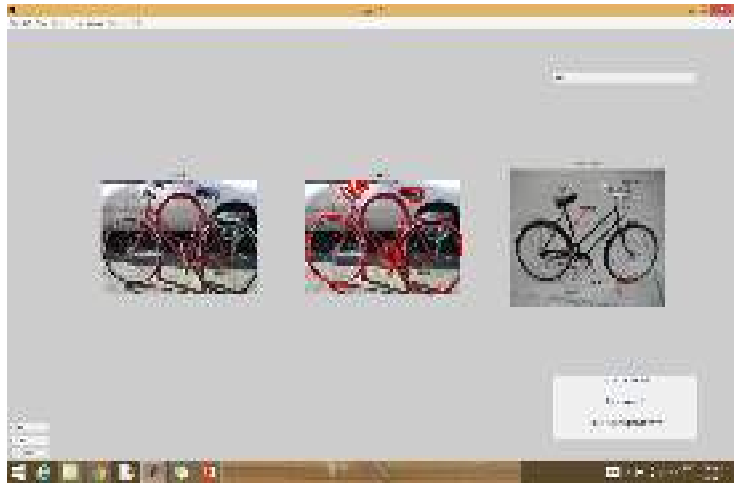

**Fig. 4.** Screenshot of annotation tool

Figure 4 shows a screenshot of our annotation tool developed in-house. For each image, the tool shows parts that have already been annotated (via renaming/simplifying of PASCAL-Parts labelings). A total of 7 annotators were given 12 categories of images (2 were given a single category of images and 5 were given 2 categories of images). Each of these categories contains 800 - 900 images. In the annotation GUI, for each object image, a drop-down list specifying the number of annotations missing in that object is provided. If the annotator feels that at least half the parts present in the drop down list can be annotated in the current image, they proceed to do so. Otherwise, the image is skipped. In this way, the annotators annotate 150 image in each category.

We provided the following annotation guidelines to annotators:
- Minimize overlap between part annotations
- If an image satisfies criteria given below, skip the image for annotation
  - Object is too small or blurry (details not noticeable enough for annotation)

- None of the object parts from the drop-down list are visible for annotation
- Handling special categories : these are categories with large number of small parts
    - `airplane` : Annotate at least 6 windows, spread across the fuselage of the plane
    - `bicycle` : Instead of individual spokes, annotate them as circular sectors (at least 6 in a wheel)
    - `cat` : Annotate whiskers as groups of 3-4 image regions on each side. Skip a group if it overlaps significantly with background.
    - `pottedplant` : Annotate leaves as groups of 3-4 image regions. Skip a group if it overlaps significantly with background.
    - `train` : Annotate at least 6 windows

Each annotator was asked to annotate 150 images. Thus, we obtain 1850 annotated images across 12 categories. We refer to this subset of PASCAL-parts with simplified/modified part annotations as PPS-12. Each object image in PPS-12 forms the basis for the construction of the image sequences in our benchmarking dataset PPSS-12, as described in the main paper.

## 3   Changing global object visibility : Part-mask based blurring

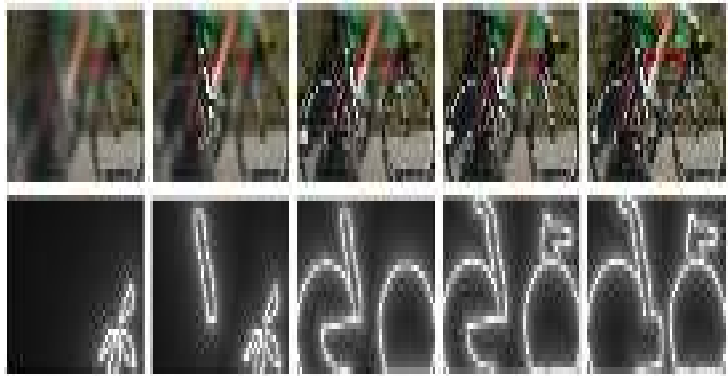

**Fig. 5.** Distance transform of the part masks (bottom row) and the final detail-reduced images (top row) for a `bicycle` image sequence.

The visibility schemes additionally have access to low-level global context from the entire image in a gist-like manner, including that from parts not yet brought into focus. In these schemes, the entire image's data is present, albeit at a low level of detail to begin with. As each part of the object is introduced, the part comes into focus. The inspiration for this scheme comes from the manner in which level of detail falls relative to the location fixated upon by a human eye. To achieve this fall-off effect, we utilize the visual-field simulation procedure described by Perry and Geisler [9]. The net effect is a blurring of the image relative to already added parts. By varying the parameters of the visual-field simulation, the level of detail in the immediate vicinity of each added part can be changed.

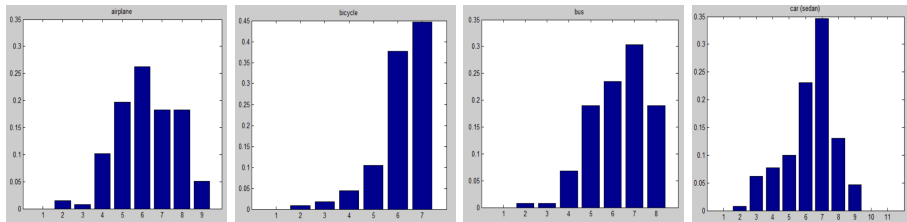

**Fig. 6.** *            **Fig. 7.** *            **Fig. 8.** *            **Fig. 9.** *
airplane            bicycle            bus            car (sedan)

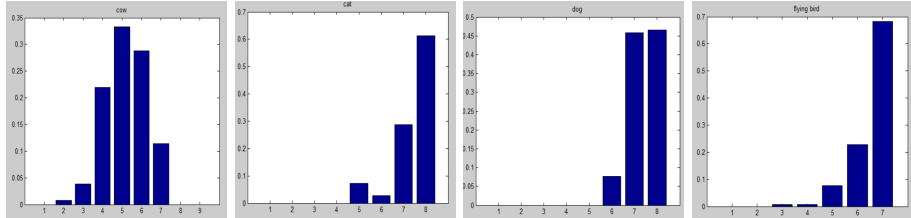

**Fig. 10.** *            **Fig. 11.** *            **Fig. 12.** *            **Fig. 13.** *
cow            cat            dog            flying bird

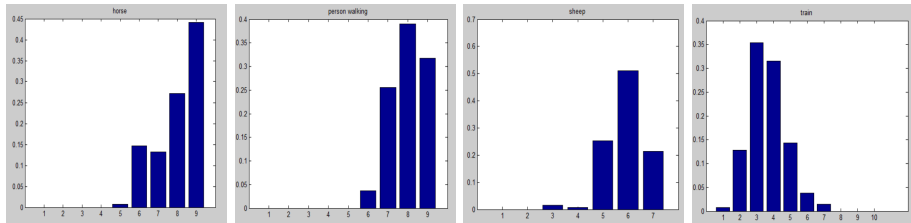

**Fig. 14.** *            **Fig. 15.** *            **Fig. 16.** *            **Fig. 17.** *
horse            person walking            sheep            train

**Fig. 18.** Part presence distribution for the 12 benchmarked object categories. In almost all the categories, the number of images with a small fraction of parts (relative to total possible part annotations) is small.

It is important to remember that this procedure is applied to each image of the part sequence. The starting point for the procedure is an image where a subset of original image parts have been added. Let us refer to the corresponding mask of parts as 'focus mask'. We compute the distance transform of this 'focus mask'. The resulting distance transform values are utilized as input to the fall-in-visual-acuity model of Perry and Geisler. Thus, the net effect is that areas further away from the part mask progressively appear blurrier (see Figure 5).

## 4    Part presence analysis

When constructing the corresponding content-modified sequence of an image $I$, the parts are added according to the order specified by a given part-scheme. Now, suppose for the object class to which $I$ belongs, there are $M$ possible parts which can be annotated in the class images. However, due to occlusions and viewpoint from which object was photographed, not all $M$ parts may be visible in $I$. Because of such missing parts, the corresponding image sequences would be of unequal length. To ensure consistency in computation of $CDS$, we do the following: Suppose, according to a part-scheme, a part has to be added to image sequence member $S_i, 1 < i \leq M$. If this part is missing in the list of annotated parts for $I$, we set $S_i = S_{i-1}$, i.e. we use the most recent part-accumulated as the current image and continue the process of adding parts.

Now, if there are a large number of missing parts in most of the images, the copying-based approach described above and the overall procedure of benchmarking would not be very meaningful. To verify this, we performed the following experiment: Suppose an object has $M$ parts and $N$ images $I_1, I_2, \ldots I_N$. Suppose $I_1$ has $m_1$ parts annotated, $I_2$ has $m_2$ parts annotated and so on. Note that $m_i \leq M$. We compute the histogram distribution of the values $\frac{m_i}{M}, i = 1, 2, \ldots N$ for each object. The manner in which these ratios are distributed tells us, on average, what fraction of total number of possible parts are actually present in the object images. The histograms for each of the 12 objects in our dataset can be seen in Figure 18. Fortunately, for most of the objects, the bin counts for smaller values (except for the category `train`) are small. Thus, there are not many images where a huge fraction of parts are missing.

# 'Part'ly first among equals: Semantic part-based benchmarking for state-of-the-art object recognition systems

Ravi Kiran Sarvadevabhatla, Shanthakumar Venkatraman, Venkatesh Babu R.

Video Analytics Lab, CDS, Indian Institute of Science
Bangalore 560012, INDIA
`ravikiran@grads.cds.iisc.ac.in`

**Abstract.** When predicted and ground-truth labels arise from different label spaces with different semantic granularity characteristics, direct comparisons are ineffective. One alternative would be to manually determine mappings between the two label spaces. However, given the subjectivity and labor involved, the resulting similarity measure becomes brittle. In such a scenario, it is appealing to characterize the 'distance' between prediction and ground-truth label strings in terms of their semantic similarity. Popular word-embedding methods such as `word2vec` and `GloVe` provide such an alternative. However, these approaches come with their own shortcomings and quirks, particularly for category names from our datasets (ILSVRC 2012, PASCAL VOC 2010). The fact that category names from ILSVRC 2012 are sourced from the English lexical database WordNet motivates the possibility of using WordNet-based similarity measures. In this work, we evaluate 9 different Wordnet-based similarity measures based on their ability to approximate manually created mappings between PASCAL categories and ILSVRC synsets. Our experiments reveal that the Wu-Palmer method, an information-content based approach best models the semantic similarity between PASCAL and ILSVRC category names.

## 1    Motivation

The setting in the main paper is as follows: We have image classifiers trained on ILSVRC-2012 dataset. Their performance is analyzed via derivative images sourced from an entirely different dataset (PASCAL VOC 2010). The datasets differ in the number of categories covered (1000 in ILSVRC v/s 20 in PASCAL) and crucially, in the semantic fine-grainedness and characterization of the category labels. PASCAL category labels are at same level of semantic granularity and are specified as a single word or a two-word phrase (e.g. `airplane, potted plant, train, dog`). On the other hand, ILSVRC labels span multiple levels of granularity (e.g. There are separate categories for the relatively closely related `dog, wolf` and `fox` categories). In addition, each ILSVRC category is a WordNet synset[1]. Typically, the first tag of the synset is coarse-grained while the other synonymous tags are more esoteric. The files `pascal-labels.txt` and `ilsvrc-labels.txt` in the supplementary material illustrate the issues mentioned above for PASCAL and ILSVRC datasets respectively.

Suppose we have an image $I$ from PASCAL VOC dataset. Suppose the ground-truth PASCAL category label is $g_I$. Given $I$ as input, suppose the label predicted by a ILSVRC-trained classifier $\mathbb{C}$ is $p_I$. Since $g_I$ and $p_I$ arise from different label spaces with different semantic granularity characteristics, a direct (string) comparison would be ineffective. Since ILSVRC labels form a larger and more diverse set than PASCAL, an alternative approach would be to group the different ILSVRC categories which map to a single PASCAL category into a single set. Suppose the ILSVRC categories which map to a PASCAL label $g$ are given by $K_g = \{i_g^1, i_g^2, \ldots\}$. The prediction $p_I$ is considered a 'correct' prediction if $p_I \in K_{g_I}$ and 'incorrect' otherwise. However, this approach has two drawbacks:

1. The grouping is done manually by a person. The grouping task can be quite laborious if the cardinality of label space is large.
2. The manual nature of grouping also introduces subjectiveness in determining the granularity of grouping (e.g. related categories of `yak` and `cow` may end up being mapped to different sets depending on the person's notion of category-level 'similarity'.). This results in a brittle non-semantic measure which cannot capture the semantic similarity that may exist (between `cow` and `yak` in the previous example).

These drawbacks naturally motivate the requirement for characterizing the 'distance' between prediction $p_I$ and ground-truth $g_I$ in terms of their semantic similarity.

## 2  Characterizing semantic similarity using word-embeddings

The larger goal for the manual mapping approach described previously is to capture the similarity between prediction and ground-truth labels. Given the shortcomings of the manual approach, we prefer to explore methods which can quantify 'semantic distance' solely based on the underlying semantics of the two words being compared.

Of late, word-embedding approaches such as `word2vec` [2] and `GloVe` [3] have gained extensive popularity for their ability to characterize semantic relationships among words. In these approaches, a word is represented as a real-valued vector in $\mathbb{R}^d$ (where $d$ typically ranges between 50 and 400). Extensive experiments using a variety of linguistic corpora have demonstrated that the semantic relationships between words can be characterized in terms of distances between the corresponding word embedding vectors.xample,

---

[1] From https://wordnet.princeton.edu/ ... WordNet [1] is a large lexical database of English. Nouns, verbs, adjectives and adverbs are grouped into sets of cognitive synonyms (synsets), each expressing a distinct concept. Synsets are interlinked by means of conceptual-semantic and lexical relations.

space2mm$|| \mathtt{w2vec}(\text{`man'}) - \mathtt{w2vec}(\text{`woman'})|| \approx || \mathtt{w2vec}(\text{`king'}) - \mathtt{w2vec}(\text{`queen'})||$
vspace2mmHowever, we discovered that these approaches come with their own shortcomings and quirks, especially when we attempt to obtain word embeddings for ILSVRC categories.

- The embedding for a multi-word category (e.g. `potted plant`) is typically handled by average pooling of the individual words (i.e. $\frac{1}{2}$( `word2vec`(`potted`) + `word2vec`(`plant`) ))
- Some ILSVRC categories (e.g. `jacamar`,`gyromitra`) are not present in the word-embedding corpora and therefore, we cannot obtain their word embeddings.
- Somewhat unintuitively, the capitalized form of a word has a different vector representation from the uncapitalized version (E.g. The cosine similarity between normalized GoogleNews corpus-based `word2vec` embeddings of `Eagle` and `eagle` is 0.7217 and not 1.0 as expected). In some cases, depending on the corpus used to generate the embedding, an embedding for the capitalized form may not even exist.
- Semantic relationships may not be preserved well between fine and coarse grained categories. For instance, for the `GloVe` embedding vectors, the cosine similarity between `dog` and `cat` is 0.80 while the similarity between `dog` and `labrador retriever` (a breed of dog) is a disturbingly low 0.09. Such dissonant results seem to exist regardless of embedding method applied (`word2vec`,`GloVe`) and word corpus used.
- Different senses of a word cannot be handled by these approaches. For instance, the word `bank` have the connotation `river bank` or it can connote a financial institution. However, both these connotations (senses) are mapped to a single representation.

**Why do these methods capture similarity semantics insufficiently and poorly?** *Firstly,* both `word2vec` and `GloVe` obtain representations and capture similarity semantics based on the co-occurrences of words at a particular abstraction of text. The two methods commonly used in the training phase of these approaches – skip-gram and CBOW – rely on document-level co-occurrences and phrase level co-occurrences respectively. Thus, an infrequent co-occurrence gets associated with a low similarity (as in the case of `dog` and `labrador retriever`). *Secondly,* it is not easy to sustain performance when the corpus is merely fine-tuned with missing words given the inherent co-occurrences the approaches rely on. For example, it is not clear how to fine-tune trained models so that `dog` and `labrador retriever` have greater similarity than `dog` and `cat`.

## 3 Characterizing semantic similarity using WordNet -based measures

For our particular situation, one set of labels (ILSVRC) are directly taken from WordNet. This property alleviates the concern of missing words which was the case for word-embedding methods. WordNet's approach groups nouns into sets of cognitive synonyms (synsets), each expressing a distinct concept. These synsets are further interlinked by means of conceptual-semantic and lexical relations. Thus, the structure of WordNet naturally captures the category hierarchy, enables different senses to be co-exist and allows semantically meaningful measures to be defined between words. The similarity measures also happen to be invariant to word capitalization. Finally, since multi-word category names are present verbatim in WordNet, they can be used by similarity measures as is without resorting to opaque word-level manipulations.

Given the above mentioned advantages, we utilize WordNet based semantic similarity measure to determine 'semantic distance' between ILSVRC-based classifier predictions $p_I$ and the ground-truth labels $g_I$ of PASCAL.

### 3.1   Selecting the similarity measure

For our experiments, we evaluate 9 similarity measures from *Wordnet::Similarity* package [4]. Broadly, similarity methods are of three categories:

1. Path-length based (*lch,wup,path*) : These similarity measures are based on path lengths between words in the WordNet graph. *lch* finds the shortest path between two words and scales that value by the maximum path length in the isa hierarchy in which they occur. *wup* finds the path length to the root node from the least common ancestor of the two words. This value is scaled by the sum of the path lengths from the individual words to the root. The measure *path* is equal to the inverse of the shortest path length between two words.

2. Information-content based (*res,lin,jcn*) : These measures are based on information content, which is a corpusbased measure of the specificity of a word. *res* computes similarity as the information content of the least common ancestor (LCA) of the two words in the Wordnet graph. The *lin* and *jcn* measures augment the information content of LCA with the sum of the information content from individual words. The *lin* measure scales the information content of the LCA by this sum, while *jcn* subtracts the information content of the LCA from this sum and then takes the inverse to convert it from a distance to a similarity measure.

3. Relatedness-based (*vector,vector_pairs,lesk*) : Each concept in WordNet is defined by a brief explanation termed gloss. The *lesk* and *vector* measures use the text of the gloss as a unique representation for the underlying concept. The *lesk* measure assigns relatedness by finding and scoring overlaps between the glosses of the two concepts, as well as concepts that are directly linked to them according to WordNet. The *vector* measure creates a cooccurrence matrix from a corpus made up of the WordNet glosses. The *vector_pairs* measure utilizes a second order co-occurrence matrix to obtain corresponding gloss vectors.

To evaluate the similarity measures, we utilize two criteria. These criteria essentially measure the extent to which the similarity scores produced by a particular measure correlate with manually created mappings between PASCAL categories and ILSVRC synsets.

**Criteria #1:** For each PASCAL category label, certain ILSVRC categories are mapped. For each mapped ILSVRC category, we determine the most similar ILSVRC category. The most similar category may or may not fall in the same mapped set. We count the number of ILSVRC categories for which the corresponding highest similarity ILSVRC category falls within the same PASCAL category mapping set. The larger the average value of this fraction for a particular similarity measure, the better the measure.

**Criteria #2:** For each PASCAL category label, we check whether the maximum similarity of label with the mapped ILSVRC tags ('intra mapped set similarity') exceeds the maximum of its similarities with other PASCAL category labels('inter mapped set similarity'). The larger the fraction of PASCAL categories for which this happens, the better the measure.

The 9 similarity measures, ranked according to the two criteria mentioned above can be seen in Tables 1 and 2.

| Similarity Measure | Criteria #1 score |
|---|---|
| *wup* | 0.945 |
| *res* | 0.938 |
| *lch* | 0.938 |
| *path* | 0.938 |
| *lesk* | 0.914 |
| *vector_pairs* | 0.914 |
| *vector* | 0.906 |
| *jcn* | 0.023 |
| *lin* | 0.023 |

**Table 1.** Similarity measures ranked by their suitability measured via Criteria #1 score - the higher the better.

| Similarity Measure | Criteria #2 score |
|---|---|
| *wup* | 0.833 |
| *res* | 0.833 |
| *lch* | 0.75 |
| *path* | 0.75 |
| *lesk* | 0.75 |
| *vector* | 0.583 |
| *vector_pairs* | 0.583 |
| *jcn* | 0.25 |
| *lin* | 0.25 |

**Table 2.** Similarity measures ranked by their suitability measured via Criteria #2 - the higher the better.

Since the measure *wup* has the highest average[2] criteria score, we utilize *wup* as our semantic similarity measure. In the main paper, we denote the similarity measure by $\mathcal{M}$ (Section 5). Therefore, $\mathcal{M} = wup$.

---

[2] Here, the average is taken over the criteria scores.

# Supplementary C : Extended set of experiments and results using Contextual Dissimilarity Score (CDS)

Ravi Kiran Sarvadevabhatla[*1], Shanthakumar Venkatraman[2] and Venkatesh Babu R.[1]

[1]Indian Institute of Science, Bangalore, INDIA
[2]Indian Institute of Technology - Hyderabad, INDIA

November 23, 2016

## Contents

## 1 Introduction

As part of the main benchmarking process, we determine the CDS for all possible combinations of classifiers, part-ordering schemes and object content schemes. This sets the stage for examining the effect of these schemes on the overall benchmarking process. For the similarity measure $\mathcal{M}$ between the category labels, we used Wu-Palmer similarity measure [5]. This measure calculates relatedness of two words using a graph-distance based method applied to WordNet [1], a standard English lexical database containing groupings of cognitively similar concepts and their interrelationships.

---

[*]ravikiran@grads.cds.iisc.ac.in

| Scheme | Context based | | Global visibility based | |
|---|---|---|---|---|
| | Intra-object | Intra-object and neighborhood | Low level of detail | Higher level of detail |
| AlexNet | 0.4499 | 0.4470 (0.6 %) | 0.4450 | 0.3803 (14.54 %) |
| GoogLeNet | 0.5264 | 0.4319 (**17.95 %**) | 0.4544 | 0.3490 (**30.20 %**) |
| NiN | 0.4788 | 0.4492 (6.18 %) | 0.4689 | 0.3882 (17.20 %) |
| VGG-19 | **0.4136** | **0.4147** (0.27 %) | **0.3628** | **0.2880** (20.62 %) |

Table 1: Benchmarking classifiers : Average median CDS across categories for different context schemes. The best CDS score for each content scheme is shown in bold. The bracketed percentages in column 2 indicate the improvement in CDS over column 1 with addition of context. The ones in column 4 indicate the improvement over column 3 when level of detail is increased. The best percentage improvement is also shown in bold. Note that smaller the CDS, better the performance.

## 1.1 Benchmarking classifiers across object content schemes

In the discussion that follows, it is important to remember that the smaller the CDS, the better the classifier's performance.

### 1.1.1 'Intra-object' context:

For the first set of experiments, we analyze CDS for the 'intra-object' context scenario. This scenario consists of object images without any of the surrounding context except that arising out of the object's parts themselves and is perhaps the most challenging scenario for a classifier. On the other hand, it is also the most appropriate since the image content is precisely confined only to the object.

Fixing the content scheme to 'intra-object', for each classifier and for each category, we compute the median CDS. We do this initially for each part scheme and subsequently average the median scores over the part schemes to obtain category-wise CDS. These category-wise scores are, in turn, averaged to obtain the CDS for each classifier. The results on a per-classifier basis can be seen in the first column ('Intra-object') of Table 1 . As expected, the median scores are generally high regardless of classifier.

### 1.1.2 'Intra-object and neighborhood' context:

We repeat the previous experiment (from Section 1.1.1) with the content-scheme now being 'intra-object and neighborhood'. In addition to object parts, contextual information from the immediate surroundings is additionally available in this scheme. We hypothesized that such information would improve performance and that is generally the case (see second column ('Intra-object and neighborhood') of Table 1).

### 1.1.3 Global object visibility:

Next, we examine the impact of visibility-based content schemes. These schemes, unlike the intra-object and/or neighborhood context ones, have additional access to low-level global context from the entire image in a gist-like manner, including that from parts not yet brought into focus. Therefore, the performance of the classifier for these schemes conveys the extent to which it utilizes the low-level global context.

Keeping the classifier fixed and content-scheme as 'low-detail', for each category, we compute the median CDS for each part scheme and average them across part schemes to obtain category-wise CDS. These are averaged in turn to obtain the CDS for the classifier. As the results in Table 1 (third column) suggest, the presence of global information, even at a low level of detail and even with minimal object-specific information, is still powerful enough to improve performance, as evidenced by the lower CDS. Increase in the level of detail (i.e. lower level of blurring) causes the results to be on predictable lines, with the overall average median CDS trending downwards (See last column of Table 1).

| Classifier | airplane | bicycle | bird | bus | car | cat | cow | dog | horse | person | sheep | train |
|---|---|---|---|---|---|---|---|---|---|---|---|---|
| Lowest-cds | GOOGLENET | VGG-19 | VGG-19 | VGG-19 | VGG-19 | VGG-19 | VGG-19 | VGG-19 | VGG-19 | GOOGLENET | ALEXNET | VGG-19 |
| Highest-cds | NiN | NiN | NiN | NiN | NiN | NiN | NiN | GOOGLENET | NiN | GOOGLENET | GOOGLENET | GOOGLENET |

Table 2: Category-wise best and worst performers (in terms of CDS) aggregated across part and content schemes.

## 1.2 Benchmarking classifiers across part schemes

For the next series of experiments, we examine the CDS keeping the part scheme fixed and aggregating across content schemes. The median CDS for the classifiers on a per-part-scheme basis can be viewed in Figure 1. In addition to the eye-fixation based part-ordering schemes, we also computed CDS for a random ordering of parts in each category. Figure 1 shows that the eye-fixation based part-schemes correspond to somewhat lower median CDS scores compared to the random counterparts. At a first glance, this result might suggest that a consistently structured part ordering (such as eye-fixation based) might not be required and *any* random part ordering might produce equivalent results in terms of CDS scores. We hypothesize that as classifiers get better (and more 'part'-aware), the gap between results of a consistently structured part scheme and random ordering will be much more. We can already see a hint of this when we specifically compare the CDS scores of top-most performers such as VGG-19 or GOOGLENET between random and non-random orderings. While we could also compare the scores by generating all the combinatorial orderings of parts, doing so would be burdensome in computation and complex in analysis.

Figure 1 also shows that the non-area-normalized eye-fixation based schemes $E_U, E_{US}$ correspond to smaller CDS in general. Although intended to guard against large parts receiving unnecessarily large importance factor, the area-based normalization, in a sense, did not matter. We can also interpret this result as demonstrating the robustness of CDS to the area-based normalization factor to some extent.

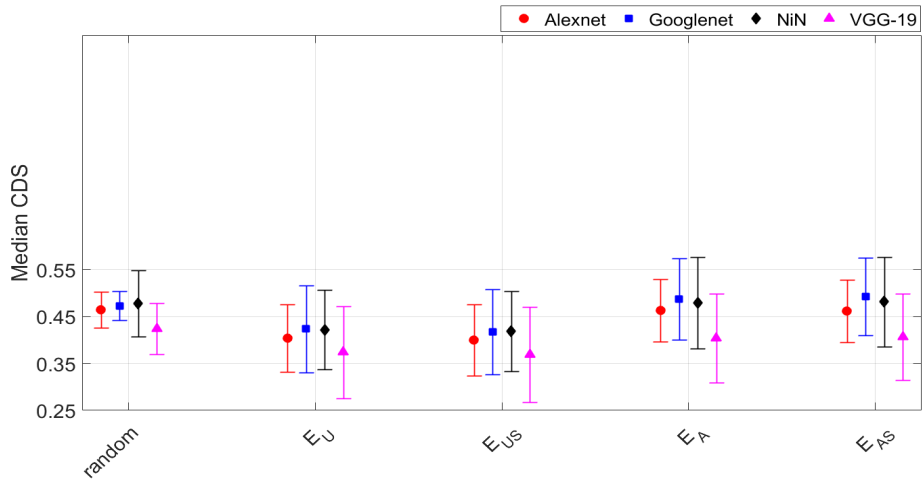

Figure 1: Part-scheme wise median CDS across content schemes for each classifier.

## 1.3 Overall performance and additional experiments

Examining the results in Table 1, it is evident that VGG-19 achieves the best performance (lowest average CDS) in general. Table 1 also shows that our benchmarking procedure contrasts the performance of almost equally well-performing classifiers (GOOGLENET, VGG-19) somewhat better than the traditional accuracy-based counterparts — the accuracy scores are closer compared to the CDS-based values in our benchmarking .

To determine which classifier exploits addition of object neighborhood-based context the most, we compute the percentage improvement in average CDS over the 'object only' (i.e. no neighborhood context) setting. As the bracketed numbers in second column of Table 1 show, GOOGLENET's performance improves the most. GOOGLENET also best exploits the increase in level of detail (fourth column of Table 1). We believe these results stems from the 'inception-style' mechanism GOOGLENET [2] uses to capture context.

To obtain a category-level perspective on the benchmarking performance, we determine the classifier that produces the lowest CDS most frequently across all combinations of part schemes and content schemes. The entries in Table 2 (top row) merely endorse the results seen earlier – VGG-19 is the best performer in general. At the other end, NIN and surprisingly, GOOGLENET have relatively higher CDS (bottom row of Table 2).

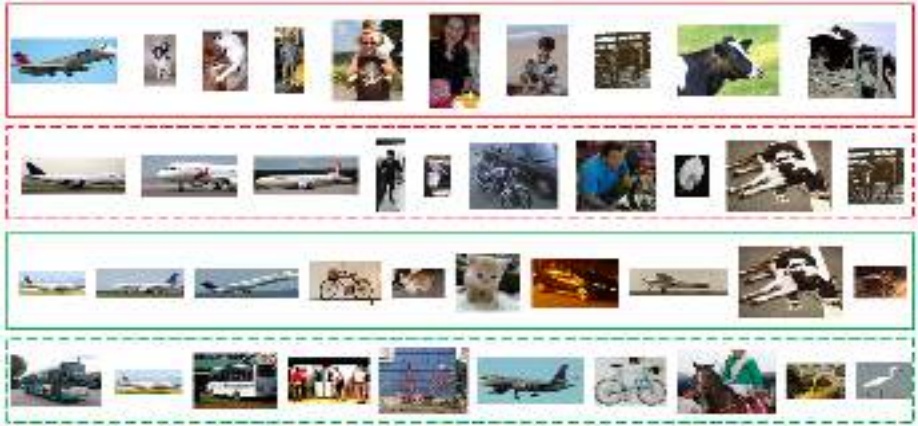

Figure 2: Quantifying the overall hardness of an image in terms of its 'semantic hardness score' (Section 1.5) : Each row of images corresponds to a fixed content scheme (from top to bottom - no context (red), neighborhood context (dashed red), high blurring (green), lower blurring (dashed green)). The images in each row are shown sorted according to their 'semantic hardness' scores – the first 3 images require only a small % of total number of parts for reliable recognition, the last 3 require the largest %. The middle 4 are the median valued images. The images are shown as their unresized, original versions.

## 1.4 Relationship between CDS and (traditional) error measures

To verify that our CDS measure provides additional information beyond the traditional top-1 error measure, we computed the correlation between CDS and the top-1 error rates across all the classifiers. For this, we determined the median error-rate and median CDS for each content scheme by averaging across the respective measures across part schemes and classifiers. Thus, we obtain two 4-dimensional vectors[1], one for median error-rate and the other for median CDS. The correlation between these two vectors was found to be close to 0 (Pearson $\rho = 0.0227, p = 0.98$ and Spearman $\rho = 0, p = 1$). This result suggests CDS measures an aspect of classifier performance distinct from the traditional top-1 measure.

## 1.5 Using CDS to characterize semantic hardness

We can also utilize CDS to quantify 'semantic hardness' of an image, i.e. in general, how hard is the task of determining the image object's identity in terms of its (image-based and prediction-label based) semantic content for a given content scheme. To do so, corresponding to each image $I$ in PPS-12, we first collect CDS of its corresponding sequences over all combinations of classifiers and part schemes. The median of all the CDS scores can be considered as the 'semantic hardness' score of $I$ for the given content scheme. We sorted all the images in PPS-12 by their 'semantic hardness' scores. For each classifier, the images with the smallest 3, median 4 and largest 3 scores can be seen in Figure 2. Across contexts, the category `cow` seems to require that most of the parts be present. On the other hand, the % of parts required seems to be lowest for `airplane` images.

---

[1]The number of content schemes is 4.

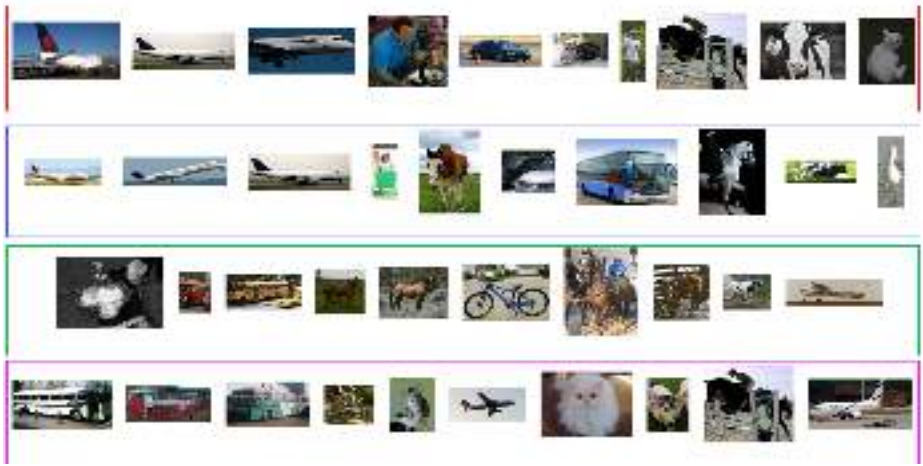

Figure 3: Quantifying the overall hardness of an image in terms of its 'semantic hardness score' (Section 1.5) : Each row of images corresponds to a classifier (from top to bottom - ALEXNET (red), GOOGLENET (blue), NIN (green), VGG-19 (pink)). The images in each row are shown sorted according to their 'semantic hardness' scores –the first 3 images require only a small % of total number of parts for reliable recognition, the last 3 require the largest %. The middle 4 are the median valued images. The images are shown as their unresized, original versions.

We repeated the above experiment, this time keeping the classifier fixed and aggregating CDS over all the content schemes and part schemes. The corresponding ranked images for each of the classifiers can be seen in Figure 3. Somewhat similar to the previous set of results, `airplane` and `bus` images can be recognized reliably even if a small number of parts are visible. On the other hand, `cow` and `horse` images require almost the entire original image presence for reliable classification.

### 1.5.1 Application: Semantic part CAPTCHAs

Since the hardness of each image for all the classifiers is known (Figure 3, the corresponding PPSS-12 sequence members can potentially be used as CAPTCHAs [3] or reCAPTCHAs [4] wherein users are asked to guess the object category. For example, suppose for a given image $I$, its semantic hardness w.r.t classifier $\mathcal{C}$ is $s_I^{\mathcal{C}}$. Since there are 2 part schemes and 4 content schemes, a total of 8 PPSS-12 sequences are associated with $I$. Suppose the length of a sequence is $N^{\mathcal{C}}$. To form a candidate pool of possible CAPTCHAs, we can pick a sequence element (partial part image) with index $i$ in the length $N$ sequence such that $1 < i < \lfloor s_I^{\mathcal{C}} \times N \rfloor$. The reasoning here is that images with high 'hardness score' typically get recognized by machines only if the total % of total parts typically exceeds $s_I^{\mathcal{C}}$. In case usage data reveals their difficulty of recognition for humans, they can be paired with an 'easy' hardness image, reCAPTCHA style. For additional rigor in picking the CAPTCHA image, we could also pick the classifier $\mathcal{C}$ with the smallest $s_I^{\mathcal{C}}$, compare it with $s' = \arg\min_{\mathcal{X}} s_I^{\mathcal{X}}$ (where $s_I^{\mathcal{X}}$ is the 'semantic hardness' of image $I$ w.r.t content scheme $\mathcal{X}$) and pick the minimum of the two values as the CAPTCHA selection threshold.

# Supplementary-D: Benchmarking using binary prediction sequences : Overview, Experiments and Analysis

Ravi Kiran Sarvadevabhatla, Shanthakumar Venkatraman, Venkatesh Babu R.

Video Analytics Lab, CDS, Indian Institute of Science
Bangalore 560012, INDIA
`ravikiran@grads.cds.iisc.ac.in`

**Abstract.** In the main paper, we proposed a novel semantic similarity measure called Contextual Dissimilarity Score (CDS). This measure relied on computation of semantic similarity between predicted and ground-truth labels for a given image. In this article, we lay out an alternative approach where we test for membership of predicted class in a set representing the ground-truth label. The membership test results in a binary similarity output for each image. We utilize such binary outputs to construct the alternate similarity measure, termed Binary Contextual Dissimilarity Score (B-CDS). In this article, we describe how B-CDS is computed. Subsequently, we analyze certain attributes of the resulting binary prediction sequences.

## 1 Introduction

In the main paper, we proposed a novel semantic similarity measure called Contextual Dissimilarity Score (CDS) (Section 5 of main paper). This measure was designed to reflect a classifier's ability to predict the target category in a semantically meaningful manner across varying visibility and contextual settings. Subsequently, we used our measure CDS and the dataset PPSS-12 to benchmark the top-performing object recognition classifiers trained on the ILSVRC-2012 dataset (Section 6 of main paper). In the follow-up discussion (Section 7 of main paper), we briefly introduced an alternative scheme for computing CDS – a traditional 'hard' $0 - 1$ binary prediction in place of 'soft' semantic similarity for each image of a PPSS-12 sequence. In this article, we shall refer to the binary prediction based similarity measure as B-CDS (B for binary).

In this article, we lay out the alternative approach for benchmarking in more detail. We first present an overview similar to the one presented in the main paper (Section 2). We then go on to describe how B-CDS is computed with this alternative approach (Section 4). Towards the end, we analyze certain attributes of the resulting prediction sequences.

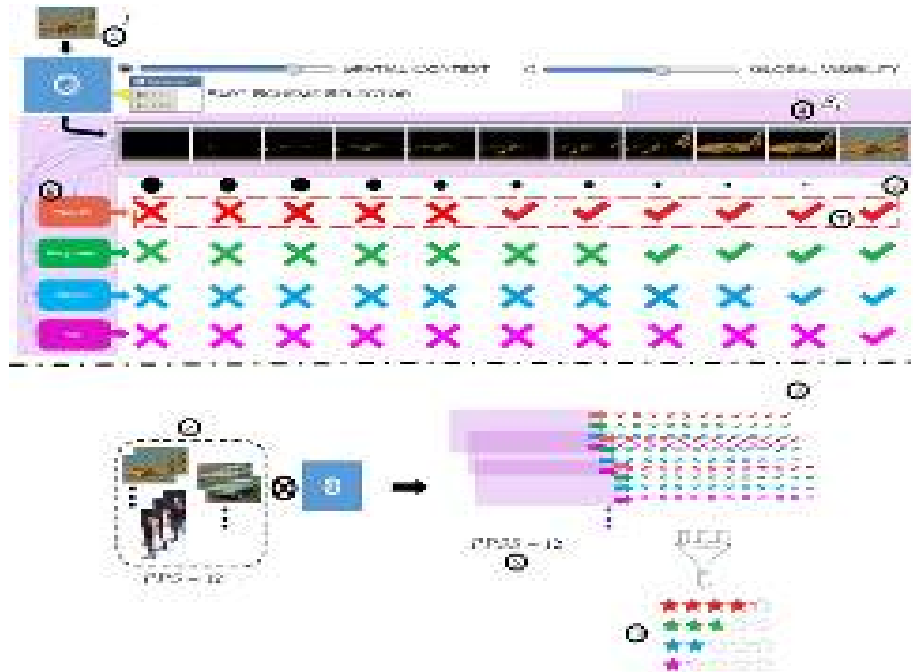

**Fig. 1.** A graphic overview of our approach (Section 2). The main processing block is shown shaded in purple background above the black dash-dotted line. ⑤ refers to collection of such blocks, each of which contains the image sequences that form our benchmarking dataset $PPSS - 12$. The ⊗ in the lower half indicates that the various global visibility schemes/spatial context schemes are applied to the base dataset $PPS - 12$ to create the sequences which form $PPSS - 12$. Also, the part-scheme, i.e. the order in which parts are incrementally added to create the image sequence $S_I$, is considered fixed for the purpose of this illustration. Best viewed in color.

## 2 Overview of the alternative approach

Figure 1 provides an overview of the alternative approach. We benchmark the top-performing [1] object classifiers trained on the 1000-class ILSVRC-2012 dataset – Alexnet [2], VGG-19 [3], NiN [4], GoogLeNet [5]. For benchmarking purposes, we first create 'PASCAL Parts Simplified (PPS)-12' – a modified, 12-category image subset (②) of PASCAL-parts [6] which in turn is a database of object images with semantic-part annotations.

For each image $I$ (①) in PPS-12 containing a reference object, we systematically vary the object's level of global visibility and its spatial context in terms of semantic object-parts (③) to create an associated sequence $S_I$ of images (④). The collection of all such sequences comprise our benchmarking dataset 'PASCAL Parts Simplified Sequences (PPSS)-12' (⑤).

Having obtained the sequences, we first fix a classifier (⑥). Passing each image sequence $S_I$ in PPSS-12 to the classifier and performing a post-prediction mapping verification (Section 3) with ground-truth set of $I$ results in a corresponding binary sequence of predictions (e.g. enclosed by dotted line near (⑦)). Each binary prediction in a sequence in associated with a weight factor such that the earlier the location of prediction, the greater its weight ((⑧)). We compute the weighted sum of predictions to arrive at the final Binary Contextual Dissimilarity Score(B-CDS). By construction, early images of the sequence contain relatively smaller amount of evidence for the reference object. Therefore, the more frequent the occurrence of correct predictions especially in the initial parts of the binary sequence, the greater the ability of the classifier to perform well in challenging conditions and demonstrate human-like performance. This is numerically characterized by a low average B-CDS for the classifier. We gather statistics on B-CDS in the image sequences (⑨) on a per-image and per-classifier basis, across object categories. These statistics enable us to benchmark the classifiers as desired (⑩).

## 3  Mapping ILSVRC predictions to PASCAL ground-truth

The setting in the main paper is as follows: We have image classifiers trained on ILSVRC-2012 dataset. Their performance is analyzed via derivative images sourced from an entirely different dataset (PASCAL VOC 2010). The datasets differ in the number of categories covered (1000 in ILSVRC v/s 20 in PASCAL) and crucially, in the semantic fine-grainedness and characterization of the category labels. PASCAL category labels are at same level of semantic granularity and are specified as a single word or a two-word phrase (e.g. `airplane`, `potted plant`, `train`, `dog`). On the other hand, ILSVRC labels span multiple levels of granularity (e.g. There are separate categories for the relatively closely related `dog`, `wolf` and `fox` categories). In addition, each ILSVRC category is a WordNet synset[1]. Typically, the first tag of the synset is coarse-grained while the other synonymous tags are more esoteric.

There are additional issues as well. ILSVRC 2012 contains around 100 categories of dogs alone. These end up mapped to a single category (`dog`) in PASCAL. In addition, a common category such as `cow` (from PASCAL) does not have even one corresponding category in ILSVRC. Therefore, we map such instances to the visually most similar categories (`water ox, Bubalus bubalis, ox`) in case of `cow`. The files `pascal-labels.txt` and `ilsvrc-labels.txt` in the supplementary material illustrate the issues mentioned above for PASCAL and ILSVRC datasets respectively.

---

[1] From https://wordnet.princeton.edu/ ...WordNet [7] is a large lexical database of English. Nouns, verbs, adjectives and adverbs are grouped into sets of cognitive synonyms (synsets), each expressing a distinct concept. Synsets are interlinked by means of conceptual-semantic and lexical relations.

**Mapping between classes in ImageNet and PASCAL-VOC:** We also need to ensure a proper mapping between classes for which the classifiers were trained (1000 classes of ImageNet) and classes from our test dataset (12 classes from PASCAL-VOC). This poses some challenges. Firstly, ImageNet classes and the PASCAL classes are not mapped one-to-one. For instance,

Suppose we have an image $I$ from PASCAL VOC dataset. Suppose the ground-truth PASCAL category label is $g_I$. Given $I$ as input, suppose the label predicted by a ILSVRC-trained classifier $\mathbb{C}$ is $p_I$. Since $g_I$ and $p_I$ arise from different label spaces with different semantic granularity characteristics, a direct (string) comparison would be ineffective. Since ILSVRC labels form a larger and more diverse set than PASCAL, an alternative approach would be to group the different ILSVRC categories which map to a single PASCAL category into a single set. Suppose the ILSVRC categories which map to a PASCAL label $g$ are given by $K_g = \{i_g^1, i_g^2, \ldots\}$. The prediction $p_I$ is considered a 'correct' prediction if $p_I \in K_{g_I}$ and 'incorrect' otherwise.

In the next section, we assume the above mentioned mapping is in place and it can be used to directly obtain the binary ('correct','incorrect') labels given $p_I$ and $g_I$. Also note that it is natural to assign a numerical value of 1 to a 'correct' prediction and 0 to an incorrect prediction.

## 4 Determining the binary Contextual Dissimilarity Score (B-CDS)

We first obtain category predictions for the entire set of images in our database. Since our analysis is on a per-classifier basis, let us fix the classifier $\mathbb{C}$ (see location ⑥ in Figure 1). For a given image $I$ from PPS-12, we first choose a part-ordering scheme and generate a sequence of images $\mathcal{S}_I$ according to this scheme (See ④ in Figure 1). We then choose an object content scheme and apply it to each of the images in the sequence. With the content-scheme applied sequence at hand, we are ready to determine B-CDS.

Let $\mathcal{S}_I = \{S_1, S_2, \ldots, S_N\}$ represent our aforementioned image sequence. Note that by construction, sequence image $S_N$ corresponds to given image $I$. Each image in the sequence is input to the classifier to obtain the corresponding class prediction. Suppose the class prediction sequence for $\mathcal{S}_I$ is given by $\{c_1, c_2, \ldots c_N\}$. We convert this sequence into a binary sequence[2] $\mathcal{L}_I = \{l_1, l_2, \ldots, l_N\}$ (Section 3).

We also note that by construction, early images of the sequences contain relatively smaller amount of evidence for the reference object. Therefore, the more frequent the occurrence of correct predictions especially in the initial parts of the sequence, the greater the ability of the classifier to perform well in challenging conditions and demonstrate human-like performance. To characterize this notion, each binary prediction $l_i$ in the sequence in associated with a weight factor $w_i = \frac{N-i+1}{N}$ such that the earlier the location of prediction, the greater its weight (See (⑤) in Figure 1)). We then compute the normalized weighted sum of binary predictions and subtract the result from 1 to arrive at the final Binary Contextual Dissimilarity Score (B-CDS).

---

[2] The colored rows containing ✗ and ✔ in Figure 1 correspond to such binary label sequences.

$$\text{B-CDS}_\text{I} = 1 - \frac{\displaystyle\sum_{i=1}^{N} l_i w_i}{\displaystyle\sum_{i=1}^{N} w_i} \tag{1}$$

The resulting B-CDS is an indicator of the part-level and contextual content required by classifier $\mathbb{C}$ to recognize the object in image $I$. In other words, a smaller average B-CDS across object categories indicates the ability of a classifier to perform well in spite of missing or poorly detailed object information.

# Supplementary E : Evaluation of semantic-similarity based benchmarking measure (CDS) and binary-prediction based benchmarking measure (B-CDS)

Ravi Kiran Sarvadevabhatla*[1], Shanthakumar Venkatraman[2] and Venkatesh Babu R.[1]

[1]Indian Institute of Science, Bangalore, INDIA
[2]Indian Institute of Technology - Hyderabad, INDIA

## Contents

*ravikiran@grads.cds.iisc.ac.in

# 1   Introduction

As part of a larger benchmarking process, we determined the CDS/B-CDS for all possible combinations of classifiers, part-ordering schemes and object content schemes. This sets the stage for not only examining the effect of these schemes on the overall benchmarking process but also for comparing CDS and B-CDS.

# 2   Benchmarking classifiers across content schemes

| Scheme | Context based | | Global visibility based | |
|---|---|---|---|---|
| | Intra-object | Intra and neighborhood | Low detail | Higher detail |
| ALEXNET | 0.4499 | 0.4470 (0.6 %) | 0.4450 | 0.3803 (14.54 %) |
| GOOGLENET | 0.5264 | 0.4319 (**17.95 %**) | 0.4544 | 0.3490 (**30.20 %**) |
| NIN | 0.4788 | 0.4492 (6.18 %) | 0.4689 | 0.3882 (17.20 %) |
| VGG-19 | **0.4136** | **0.4147** (0.27 %) | **0.3628** | **0.2880** (20.62 %) |

Table 1: Average median CDS across categories for different context schemes.

| Scheme | Context based | | Global visibility based | |
|---|---|---|---|---|
| | Intra-object | Intra and neighborhood | Low detail | Higher detail |
| ALEXNET | 0.9987 | 0.9975 (-0.1 %) | 0.9905 | 0.9821 (0.8 %) |
| GOOGLENET | 0.9963 | 0.9886 (**0.78 %**) | 0.9863 | 0.9588 (**2.79 %**) |
| NIN | 0.9979 | 0.9979 (0 %) | 0.9960 | 0.9788 (1.73 %) |
| VGG-19 | **0.9933** | **0.9886** (0.47 %) | **0.9753** | **0.9554** (2.04 %) |

Table 2: Average median B-CDS across categories for different context schemes.

As can be seen, the score values for B-CDS are extremely high compared to the scores of CDS. One reason is that for any given prediction of a PPSS-12 sequence image, the binary prediction value forms a lower bound to similarity measure when an incorrect prediction takes place, but it forms an upper bound when a correct prediction takes place. Consider an incorrectly classified image from PPS-12. Our analysis reveals that for most of the misclassified images, most of the corresponding PPSS-12 sequence are misclassified too, leading to a low weighted score. Therefore, the corresponding B-CDS score is high.

# 3   Benchmarking classifiers across part schemes

For the next series of experiments, we examined the CDS keeping the part scheme fixed and aggregating across content schemes. The median CDS/B-CDS for the classifiers on a per-part-scheme basis can be viewed in Figures 1 and 2.

# 4   Overall performance and additional experiments

To obtain a category-level perspective on the benchmarking performance, we determine the classifier that produces the lowest CDS most frequently across all combinations of part schemes and content schemes. The entries in Table 3 (top row) merely endorse the results seen earlier – VGG-19 is the best performer in general. At the other end, NiN and surprisingly, GoogLeNet have relatively higher CDS (bottom row of Table 3).

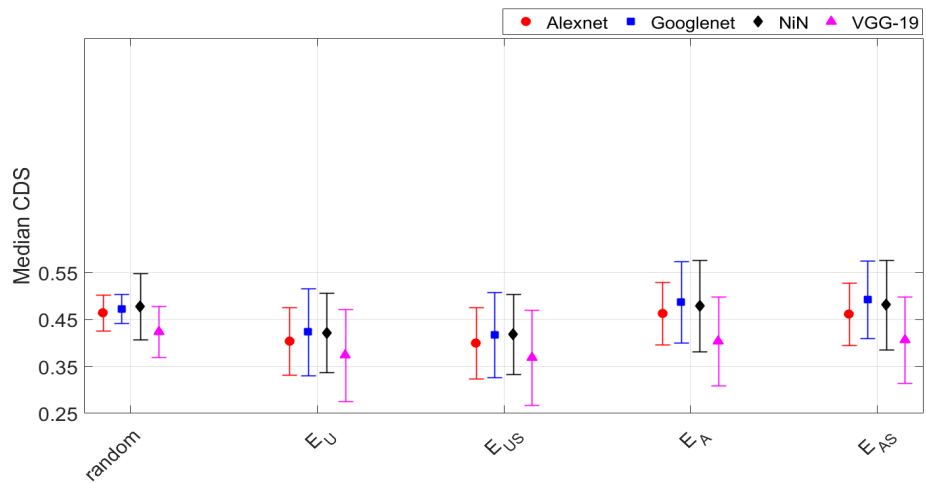

Figure 1: Part-scheme wise median CDS across content schemes for each classifier.

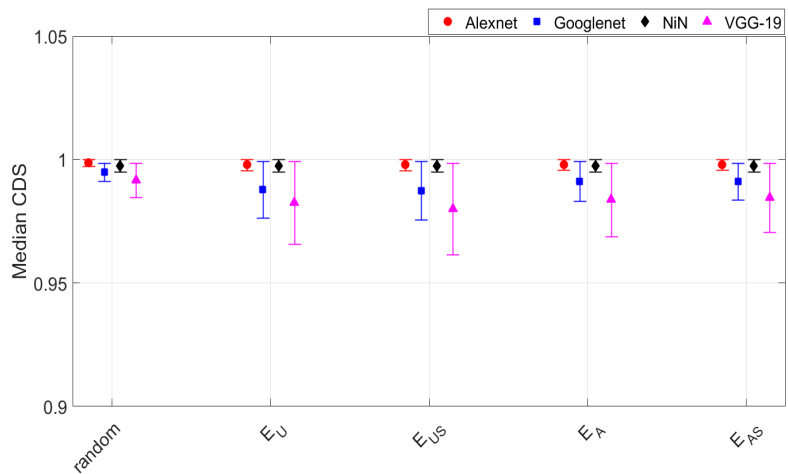

Figure 2: Part-scheme wise median B-CDS across content schemes for each classifier.

Results for a similar experiment using B-CDS as the measure can be seen in Table 4.

| Classifier | airplane | bicycle | bird | bus | car | cat | cow | dog | horse | person | sheep | train |
|---|---|---|---|---|---|---|---|---|---|---|---|---|
| Lowest-cds | GoogLeNet | VGG-19 | VGG-19 | VGG-19 | VGG-19 | VGG-19 | VGG-19 | VGG-19 | VGG-19 | VGG-19 | AlexNet | VGG-19 |
| Highest-cds | NiN | NiN | NiN | NiN | NiN | NiN | NiN | GoogLeNet | NiN | GoogLeNet | GoogLeNet | GoogLeNet |

Table 3: Category-wise best and worst performers (in terms of CDS) aggregated across part and content schemes.

| Classifier | airplane | bicycle | bird | bus | car | cat | cow | dog | horse | person | sheep | train |
|---|---|---|---|---|---|---|---|---|---|---|---|---|
| Lowest-b-cds | VGG-19 | VGG-19 | VGG-19 | VGG-19 | VGG-19 | VGG-19 | GoogLeNet | VGG-19 | VGG-19 | AlexNet | VGG-19 | GoogLeNet |
| Highest-b-cds | NiN | AlexNet | NiN | NiN | NiN | AlexNet | AlexNet | AlexNet | NiN | AlexNet | VGG-19 | AlexNet |

Table 4: Category-wise best and worst performers (in terms of B-CDS) aggregated across part and content schemes.

# 5 Relationship between CDS/B-CDS and (traditional) error measures

To verify that our CDS measure provides additional information beyond the traditional top-1 error measure, we computed the correlation between CDS and the top-1 error rates across all the classifiers. For this, we determined the median error-rate and median CDS for each content scheme by averaging across the respective measures across part schemes and classifiers. Thus, we obtain two 4-dimensional vectors[1], one for median error-rate and the other for median CDS. The correlation between these two vectors was found to be close to 0 (Pearson $\rho = 0.0227, p = 0.98$ and Spearman $\rho = 0, p = 1$). This result suggests CDS measures an aspect of classifier performance distinct from the traditional top-1 measure.

For the same experiment, when performed using the B-CDS values, the correlation between error-rate and B-CDS values was found to be non-trivially far from 0 (Pearson $\rho = -0.33204$) but with $p = 0.6680$ thus indicating a high degree of uncertainty regarding any interpretation of the $\rho$ value. Therefore, it not clear to what extent the information provided by B-CDS overlaps with that provided by the traditional top-1 measure.

---

[1] The number of content schemes is 4.

# A  Examining the viability of the overall sequence-weighting scheme used in B-CDS/CDS

## A.1  A brief review of CDS/B-CDS

For a given image $I$ from our PPS-12 dataset, we first choose a part-ordering scheme and generate a sequence of images $\mathcal{S}_I$ according to this scheme. We then choose an object content scheme and apply it to each of the images in the sequence. With the content-scheme applied sequence at hand, we are ready to determine the Contextual Dissimilarity Score $CDS_I$ /$B - CDS_I$ for image $I$.

### A.1.1  CDS

Let $\mathcal{S}_I = \{S_1, S_2, \ldots, S_N\}$ represent our aforementioned image sequence. Note that by construction, sequence image $S_N$ corresponds to given image $I$. Since our analysis is on a per-classifier basis, let us fix the classifier $\mathbb{C}$. Each image in sequence $S_j \in \mathcal{S}_I$ is input to the classifier to obtain the corresponding class prediction label $c_j$. Suppose the ground-truth label for $S_j$ is $g_j$. In our case, $c_j$ and $g_j$ are drawn from two different label spaces (Imagenet-based and PASCAL-based) with varying levels of semantic granularity and therefore, an exact literal match may not be possible. Therefore, we utilize a semantic similarity measure $\mathcal{M}$ which provides a $[0, 1]$-normalized score $x_j$ reflecting the semantic similarity between $c_j$ and $g_j$ (i.e. $x_j = \mathcal{M}(c_j, g_j)$). Thus, we obtain a sequence of normalized scores $\mathcal{X}_I = \{x_1, x_2, \ldots, x_N\}$. Also note that by construction, early images of the sequences contain relatively smaller amount of evidence for the reference object. Therefore, the higher the similarity score in the initial parts of the sequence, the greater the ability of the classifier to perform well in challenging conditions and demonstrate human-like performance. To characterize this notion, each similarity score $x_j$ in the sequence in associated with a weight factor $w_j = \frac{N-j+1}{N}$ such that the earlier the location, the greater its weight. We then compute the normalized weighted sum of similarity scores and subtract the result from 1 to arrive at the final score for image $I$ ($CDS_I$).

$$CDS_I = 1 - \frac{\displaystyle\sum_{j=1}^{N} x_j w_j}{\displaystyle\sum_{j=1}^{N} w_j} \tag{1}$$

### A.1.2  B-CDS

Let $\mathcal{S}_I = \{S_1, S_2, \ldots, S_N\}$ represent our aforementioned image sequence. Note that by construction, sequence image $S_N$ corresponds to given image $I$. Each image in the sequence is input to the classifier to obtain the corresponding class prediction. Suppose the class prediction sequence for $\mathcal{S}_I$ is given by $\{c_1, c_2, \ldots c_N\}$. We convert this sequence into a binary sequence $\mathcal{L}_I = \{l_1, l_2, \ldots, l_N\}$. We also note that by construction, early images of the sequences contain relatively smaller amount of evidence for the reference object. Therefore, the more frequent the occurrence of correct predictions especially in the initial parts of the sequence, the greater the ability of the classifier to perform well in challenging conditions and demonstrate human-like performance. To characterize this notion, each binary prediction $l_i$ in the sequence in associated with a weight factor $w_j = \frac{N-j+1}{N}$ such that the earlier the location of prediction, the greater its weight. We then compute the normalized weighted sum of binary predictions and subtract the result from 1 to arrive at the final Binary Contextual Dissimilarity Score (B-CDS).

$$\text{B-CDS}_\text{I} = 1 - \frac{\sum\limits_{j=1}^{N} l_j w_j}{\sum\limits_{j=1}^{N} w_j} \qquad (2)$$

## A.2   CL-ratio experiment

In this section, we shall restrict our focus to B-CDS. The approach put forth in this section is applicable to CDS with some modifications.

In our measurement schemes B-CDS, we make a crucial assumption in the form of the weight factor $w_i$. Suppose for a given classifier $\mathcal{C}$, the binary prediction sequence for image $I \in PPS - 12$ is $\mathcal{L}_\mathcal{I}$ (Section A.1.2). Suppose the sequence has the binary pattern that early positions of the sequence have a few 1s but later positions (including the last) are 0 (E.g. $[0, 0, 1, 1, 1, 1, 0, 0, 0, 0, 0, 0, ]$). The binary pattern corresponds to a situation where the classifier, for some reason, fails to correctly recognize $I$ even when additional part-based evidence is successively added. Our weighing scheme reflects our fundamental assumption that early 1s correspond to a better classifier. However, for sequences described previously (early 1s, later 0s), the weighing scheme assigns an incorrect (higher) performance rating to the classifier. To determine the extent to which such sequences exist, we performed the following experiment.

Now, consider a $N$-length sequence $\mathcal{L}_I^N = \{l_1, l_2, \ldots, l_N\}$. For each position $i, 1 <= i <= N$ in the sequence,we identify positions $j$ such that $l_{j-1} = 0, l_j = 1$. Each such position has the possibility of being the starting member of a subsequence consisting solely of correct predictions (1s). Suppose the length of the subsequence starting at $l_j$ is $s_j$. Let the length of total remaining original sequence relative to position $j$ and inclusive of $j$ is $t_j = N - j + 1$. Let us define $CL_j = \frac{s_j}{t_j}$. $CL_j$ reflects the extent to which the contiguous sequence of correct predictions continues until the final, full image in the corresponding PPSS-12 subsequence.

We first sort the sequences occurring across all images, across all categories and part schemes (i.e. whole of PPSS-12) in the increasing order of their length. Suppose the maximum sequence length is $M$. Thus, there are a maximum of $M$ possible indices for a PPSS-12 sequence. For each position $i, 1 \leq i \leq M$ over all the sequences, we determine the corresponding $CL_i$s. Thus, we obtain a set of such ratio values for each position. We form a 10-bin histogram for each position.

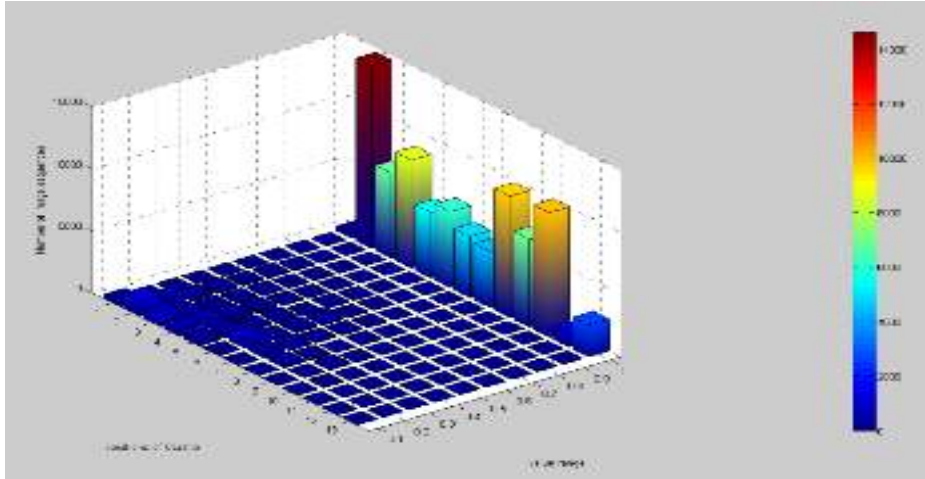

Figure 3: The sequence index positions are on x-axis, the bin centers of the 10-bin histogram are on y-axis and the histogram counts are on z-axis. As can be seen, the number of bin entries in the bottom left corner are quite small, thus indicating that our weighing scheme is valid for the most part across all categories, classifiers and part schemes.

The unwelcome situation would be presence of a large number of small $CL_i$ ratios, especially for small values of $i$ since that would mean a significant number of 'early 1s and later 0s' sequences. As can be seen from Figure 3, we fortunately do not have such a situation across the part sequences. Therefore, our benchmarking procedure and resulting scores are quite valid.